\documentclass[pdflatex,sn-mathphys-num]{sn-jnl}

\usepackage{indentfirst} 
\usepackage{graphicx}
\usepackage{multirow}
\usepackage{amsmath,amssymb,amsthm,mathtools,bm,siunitx,float,amsfonts}
\usepackage{mathrsfs}
\usepackage[title]{appendix}
\usepackage{xcolor}
\usepackage{textcomp}
\usepackage{manyfoot}
\usepackage{booktabs}
\usepackage{algorithm}
\usepackage{algorithmicx}
\usepackage{subcaption}
\usepackage{algpseudocode}
\usepackage{listings}
\usepackage{subcaption}
\usepackage{rotating} 
\definecolor{blue}{RGB}{0, 60, 120}

\theoremstyle{thmstyleone}%
\newtheorem{theorem}{Theorem}
\theoremstyle{thmstyletwo}%
\newtheorem{remark}{Remark}%

\theoremstyle{thmstylethree}%

\begin{document}

\title[SpectralTrain: A Universal Framework for Hyperspectral Image Classification]{SpectralTrain: A Universal Framework for Hyperspectral Image Classification}


\author[1,2]{\fnm{Meihua} \sur{Zhou}}
\email{mhzhou0412@gmail.com, zhoumeihua25@mails.ucas.ac.cn}
\equalcont{Meihua Zhou and Xinyu Tong contributed equally to this work.}

\author[1]{\fnm{Liping} \sur{Yu}}
\email{ylp1992247079@gmail.com}

\author*[2]{\fnm{Xinyu} \sur{Tong}}
\email{tongxinyu25@mails.ucas.ac.cn}
\equalcont{Meihua Zhou and Xinyu Tong contributed equally to this work.}

\author[3]{\fnm{Wai Kin} \sur{Fung}}
\email{1155208542@link.cuhk.edu.hk}

\author[4]{\fnm{Ruiguo} \sur{Hu}}
\email{hrg010923@163.com}

\author[2]{\fnm{Jiarui} \sur{Zhao}}
\email{zhaojiarui@binn.cas.cn}

\author*[1]{\fnm{Nan} \sur{Wan}}
\email{wannan@wnmc.edu.cn}

\affil[1]{\orgdiv{School of Medical Information}, \orgname{Wannan Medical University}, \orgaddress{\city{Wuhu}, \country{China}}}

\affil[2]{\orgname{University of Chinese Academy of Sciences}, \orgaddress{\city{Beijing}, \country{China}}}

\affil[3]{\orgname{The Chinese University of Hong Kong}, \orgaddress{\city{Hong Kong}, \country{China}}}

\affil[4]{\orgname{College of William \& Mary}, \orgaddress{\city{Williamsburg}, \state{Virginia}, \country{United States}}}

\abstract{
Hyperspectral imaging provides high-dimensional spectral sensing for industrial inspection, environmental monitoring, agricultural assessment, and Earth observation, yet its deployment is often constrained by expensive training, limited labeled samples, and large host--device data movement. Most existing hyperspectral image classification studies address these constraints by redesigning model architectures, which improves accuracy but leaves the training pipeline largely unchanged. This study introduces SpectralTrain, an architecture-agnostic fast-learning framework for hyperspectral image classification. The core idea is to organize training as a spectral curriculum. Early stages use principal component analysis to form compact, information-preserving spectral views, and later stages progressively restore full spectral complexity as optimization stabilizes. This design reduces early-epoch computation and data transfer while preserving the ordered spectral structure that is essential for material discrimination. SpectralTrain is independent of backbone, optimizer, and loss function, making it compatible with convolutional, 3D spectral, and transformer-based models. Experiments on Indian Pines, Salinas-A, and CloudPatch-7 show consistent 2--7$\times$ training acceleration with small accuracy variation across datasets and backbones. The cloud classification study further demonstrates its relevance to climate-oriented hyperspectral sensing. These results indicate that training-schedule optimization is an effective and deployable complement to architectural innovation for industrial hyperspectral intelligence.
}


\maketitle

\section{Introduction}\label{sec1}

Hyperspectral imaging (HSI) is an important sensing modality for industrial artificial intelligence because it records dense spectral responses over hundreds of contiguous bands and enables fine-grained material, land-cover, atmospheric, and environmental discrimination \cite{1,2,3}. It has been applied in agricultural assessment, Earth observation, environmental monitoring, cloud and atmosphere analysis, and inspection-oriented intelligent sensing \cite{22,23}. Compared with RGB imaging, HSI provides richer physical and chemical information, but this advantage also introduces a typical industrial AI bottleneck. Hyperspectral data are high-dimensional, labels are often expensive, sensor responses vary across platforms, and training pipelines are constrained by memory, computation, and host--device bandwidth \cite{4,5}. These constraints make efficient learning as important as model accuracy for practical hyperspectral intelligence.

Existing HSI classification studies have advanced spectral--spatial representation through convolutional, 3D spectral, attention, transformer, generative, and capsule-based models \cite{1,5,27,28,29,30,31}, while related work on super-resolution, fusion, computational imaging, restoration, and spatial--spectral dictionary learning emphasizes preserving spectral structure during representation improvement \cite{2,6,7,42}. However, efficiency is still pursued mainly through architectural design, and training usually follows a full-spectrum, full-resolution loop. This is inefficient because early epochs often need only dominant spectral components for coarse class separation, whereas full spectral complexity becomes more useful after optimization stabilizes. Starting from complete hyperspectral cubes therefore wastes computation and data movement before fine spectral cues are fully exploited.

Designing an efficient training schedule for HSI is challenging because hyperspectral data differ fundamentally from RGB images. HSI cubes have contiguous spectral ordering, sensor-dependent band responses, and class-discriminative cues that may concentrate in narrow wavelength intervals \cite{3,4,44}. Uniform band dropping, random spectral truncation, or spatial low-frequency cropping can remove informative bands, distort spectral signatures, or mix neighboring spectra across class boundaries \cite{43,45,46}. Therefore, an HSI-oriented curriculum should reduce early-stage training cost while preserving dominant spectral structure and avoiding irreversible degradation of per-pixel spectral information.

Motivated by curriculum learning and efficient visual backbone training \cite{8,18}, we propose \textit{SpectralTrain}, a universal training framework for hyperspectral image classification. SpectralTrain begins with compact principal component analysis (PCA) projections that retain dominant spectral energy and reduce early training cost \cite{47}. It then progressively increases spectral complexity until the full hyperspectral representation is restored. This progressive schedule acts as a continuation process from low-complexity spectral views to full-spectrum learning, allowing the optimizer to first learn stable coarse separations and then refine class boundaries with richer spectral evidence. SpectralTrain is designed as a drop-in framework for industrial hyperspectral learning. It does not modify the model architecture, optimizer, or loss function, and can be applied to classical convolutional networks, 3D spectral networks, and transformer-based models \cite{24,25,26,28,29,30}. Since PCA is applied before GPU upload, early stages reduce both computation and data transfer, which is especially relevant for resource-constrained training and edge-oriented hyperspectral systems.

We evaluate SpectralTrain on Indian Pines, Salinas-A, and CloudPatch-7. These datasets cover different spatial scales, spectral characteristics, and application settings, including a climate-related cloud classification scenario \cite{23}. Across representative backbones, SpectralTrain achieves 2--7$\times$ training speedups with small accuracy variation, indicating that spectral curriculum learning can substantially reduce training cost while maintaining competitive classification performance. These results suggest that efficient training schedules provide a practical route toward scalable hyperspectral intelligence in industrial and environmental sensing.

The main contributions of this work are summarized as follows:
\begin{itemize}
\item We identify training efficiency as a central but underexplored problem in hyperspectral image classification under industrial AI constraints, including high-dimensional sensing, limited labels, sensor heterogeneity, and bandwidth-bound training.

\item We propose SpectralTrain, an architecture-agnostic spectral curriculum that starts from PCA-compressed spectral views and progressively restores full spectral complexity, enabling efficient low-to-high spectral learning without changing model architectures, optimizers, or loss functions.

\item We validate SpectralTrain on Indian Pines, Salinas-A, and CloudPatch-7, demonstrating 2--7$\times$ training acceleration with small accuracy variation across datasets and backbones, and showing its potential for climate-oriented hyperspectral sensing.
\end{itemize} 

\section{Related Work}\label{sec2}
\subsection{Training Strategy in Computer Vision}
Recent advances in deep learning have highlighted the strength of pretraining frameworks that decouple representation learning from task-specific optimization. In natural language processing, models such as GPT \cite{9} and BERT \cite{10} pretrain on large corpora, while in computer vision, MAE \cite{11} and DINO \cite{12} learn transferable visual features, yielding notable gains in data efficiency and downstream performance \cite{13}. However, these approaches typically require substantial compute and extensive data, which limits practicality for hyperspectral image (HSI) \emph{classification}, where labels are scarce and the spectral axis dominates memory and I/O.

In parallel, task-specific training frameworks improve schedules, inputs, and policies under domain or resource constraints \cite{14,15}, including self-supervised iterative training for 3D point cloud denoising \cite{16} and reinforcement-learning-based configuration for resource-limited federated learning \cite{17}. EfficientTrain++ \cite{8} further shows that a resolution-based curriculum can balance visual richness and training cost. However, HSI classification requires a spectral-aware curriculum because HSI cubes contain contiguous sensor-induced spectra and band-localized discriminative cues, making resolution-only schedules insufficient and potentially harmful to informative bands or early optimization.

\subsection{Curriculum Learning}
Curriculum learning (CL) organizes the training process by presenting easier elements before harder ones \cite{18}. Representative variants integrate fuzzy-clustered curricula for semi-supervised classification \cite{19} and design multi-task curricula for weakly supervised learning \cite{20}. Self-paced and difficulty-aware schedules further use optimization feedback or uncertainty estimation to select and weight samples, improving robustness to label noise and class imbalance \cite{21}. These methods typically operate along sample-, label-, or objective-level difficulty and have proven effective for convergence and generalization.

Beyond sample ordering, curricula can also control the input signal itself (e.g., image resolution or augmentation strength) to shape the optimization landscape. However, most such designs assume RGB image statistics. In hyperspectral imaging, information is distributed along a contiguous spectral axis with instrument-specific responses, and class-discriminative cues are often band-localized rather than uniformly spread. Directly borrowing RGB-oriented curricula can therefore discard informative bands or destabilize early optimization.

We position SpectralTrain as a curriculum grounded in the spectral axis: training starts from an information-preserving principal component analysis (PCA) compression of spectra and progressively restores bands as learning stabilizes. This spectral curriculum preserves task-relevant components while reducing early-epoch compute and I/O. It is orthogonal to spatial or semantic curricula and is compatible with a wide range of architectures and loss functions.

\subsection{Hyperspectral Imaging for Climate and Environment}
Hyperspectral imaging has demonstrated considerable potential in climate and environmental monitoring \cite{22}. By capturing per-pixel reflectance across contiguous wavelengths, it supports aerosol and gas detection, vegetation stress analysis, and water quality assessment through material-specific spectral signatures. These capabilities provide a fine-grained basis for environmental state estimation beyond broadband sensors. Despite this potential, adoption in meteorology and climate prediction remains comparatively limited. We study cloud-type classification as a practical entry point: the CloudPatch-7 dataset \cite{23} assembles hyperspectral cloud patches to analyze spectral and spatial patterns relevant to early-stage weather recognition. Spectral cues associated with particle size, phase (ice versus liquid water), and optical thickness complement spatial texture, offering a discriminative basis for automatic cloud typing.

Operational scenarios impose tight compute and memory budgets, while
hyperspectral datacubes are large and bandwidth-bound. These constraints
motivate efficient training frameworks that respect spectral structure.
Existing hyperspectral image classification studies commonly retain
full spectral complexity throughout training. Organizing training to begin
with information-preserving spectral compression and progressively increase
spectral complexity provides a practical means of reducing early-stage cost.

\section{Method}\label{sec3}
SpectralTrain differs from prior hyperspectral studies that mainly improve specific model architectures. Instead of redesigning the backbone, it optimizes the training pipeline through PCA-based spectral compression and curriculum learning. Traditional hyperspectral research generally concentrates on optimizing a particular model’s performance (e.g., fine-tuning feature extractors or classifiers). In contrast, SpectralTrain introduces a universal training framework that combines principal component analysis (PCA) based image-level spectral downsampling with a curriculum learning (CL) strategy, thereby redefining the training pipeline.

\textcolor{blue}{The theoretical analysis below separates exact properties
of the spectral curriculum from an explanatory optimization analysis under
explicit regularity assumptions. The exact results characterize the nested
PCA representation, the discarded spectral energy, and the
input-dependent computational cost. The optimization result explains the
continuation mechanism through a smooth strongly convex surrogate and does
not serve as a global convergence or accuracy guarantee for the deep
non-convex models used in the experiments.}

\begin{figure}[h]
  \centering
  \includegraphics[width=\textwidth]{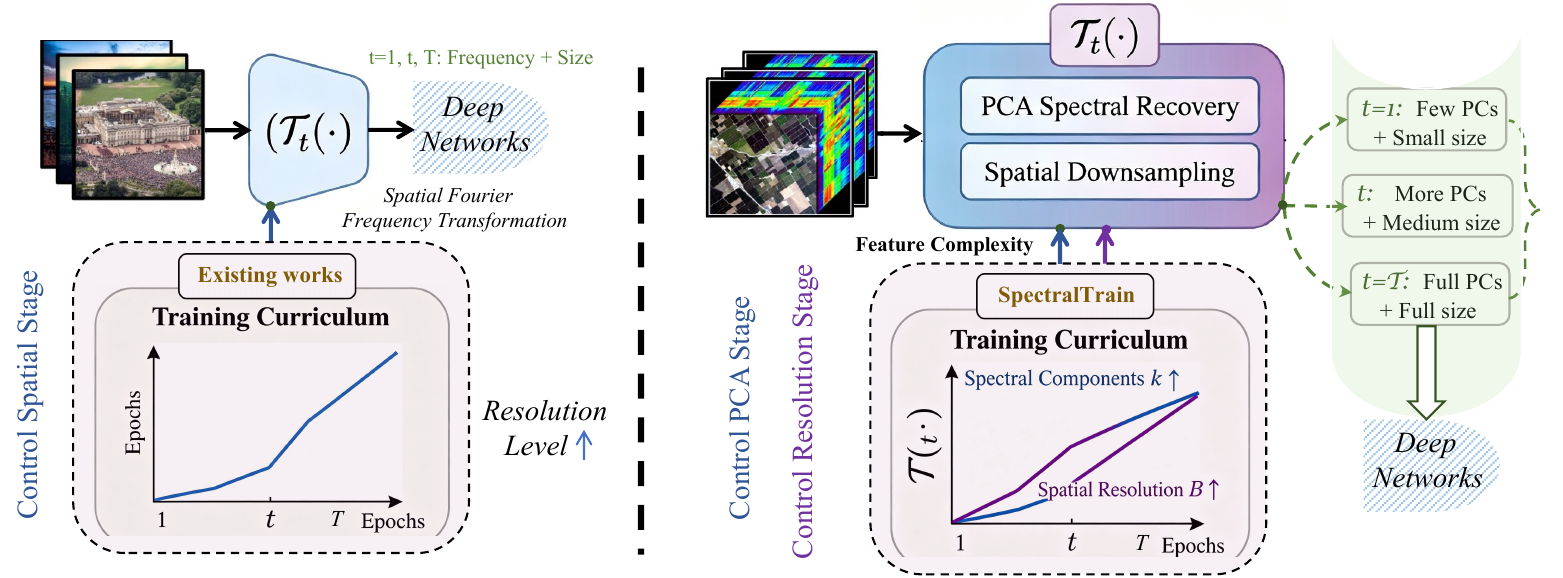}
  \caption{Overview of the proposed SpectralTrain framework. 
\textbf{Left subplot} illustrates the generalized curriculum learning (CL) pipeline of existing methods for RGB natural images. A single \(\mathcal{T}_t(\cdot)\) with spatial Fourier frequency transformation is controlled by the Training Curriculum scheduler, which only regulates the spatial resolution level as training epochs increase from \(1\) to \(T\). During training, \(\mathcal{T}_t(\cdot)\) outputs samples with gradually elevated spatial frequency and image size (\(t=1, t, T\): Low/Mid/Full Frequency + Small/Medium/Full size) and feeds them into deep networks. Only one single spatial control dimension is involved in this paradigm. \textbf{Right subplot} shows our extended spatio-spectral generalized CL named SpectralTrain tailored for hyperspectral image (HSI) cubes. The transformation module \(\mathcal{T}_t(\cdot)\) contains two parallel branches: PCA Spectral Recovery and Spatial Downsampling. Correspondingly, the Training Curriculum scheduler provides two independent control signals: Control PCA Stage for tuning the number of retained spectral principal components \(k\), and Control Resolution Stage for adjusting spatial resolution \(B\). As epochs grow, \(\mathcal{T}_t(\cdot)\) progressively generates HSI samples with rising spectral component quantity and spatial size: \(t=1\) (Few PCs + Small size), intermediate epoch \(t\) (More PCs + Medium size), and final epoch \(T\) (Full PCs + Full size), before sending complex spatio-spectral data to deep networks. \textbf{Different from RGB-only single spatial curriculum, SpectralTrain synchronously exposes richer spectral and spatial features via dual-dimensional soft-selection transformation, forming a joint spatio-spectral generalized curriculum learning strategy for hyperspectral tasks.}}
  \label{fig:overview}
\end{figure}

This framework is not confined to a single backbone; it is applicable to convolutional neural networks (CNNs), 3D spectral networks, and transformer-based models. PCA reduces input spectral dimension while preserving dominant information; CL organizes training from easy to hard by gradually restoring spectral complexity and increasing spatial resolution. A schematic overview is shown in Fig.~\ref{fig:overview}.

\subsection{Computational Constraints and Optimization Objectives}\label{subsec:constraints}

We aim to maximize prediction accuracy under a fixed compute budget. Let total training epochs be
\(
T=\beta T_0
\),
where \(T_0\) is the baseline (full-cost) training epochs and \(0<\beta<1\) is a pre-defined saving ratio. Training is divided into \(N\) curriculum stages. Each stage uses a spatial resolution \(B_i\) and retains \(k_i\) principal components of the original spectral dimension \(D\) (\(k_i \le D\)), with \(B_i\) and \(k_i\) increasing across stages.

A piecewise-continuous learning-rate scheduler \(\alpha_{\mathrm{lr}}(t_1{:}t_2)\) is adopted; the interval \(((i-1)/N,\, i/N]\) is reserved for stage \(i\). To balance compute across stages, we scale the number of optimization steps by spatial FLOPs:
\begin{equation}
\mathrm{Steps}_i
=
\left\lfloor
\frac{T}{N}\cdot \left(\frac{B_0}{B_i}\right)^{2}
\right\rfloor,
\quad\text{since per-step FLOPs }\propto B^{2}\text{ for standard convolutions.}
\label{eq:steps}
\end{equation}
where \(B_0\) is the original spatial size. For spectral compression, the transfer/computation reduction factor is
\begin{equation}
\mathrm{Compression\ Ratio}=\frac{D}{k_i},
\label{eq:cr}
\end{equation}
which upper-bounds the saving in host–device transfer; activation memory also depends on architecture and depth when PCA is applied prior to upload.

\subsection{Training Schedule and Stage Switching}\label{subsec:schedule}

\begin{figure}[h]
  \centering
  \includegraphics[width=\textwidth]{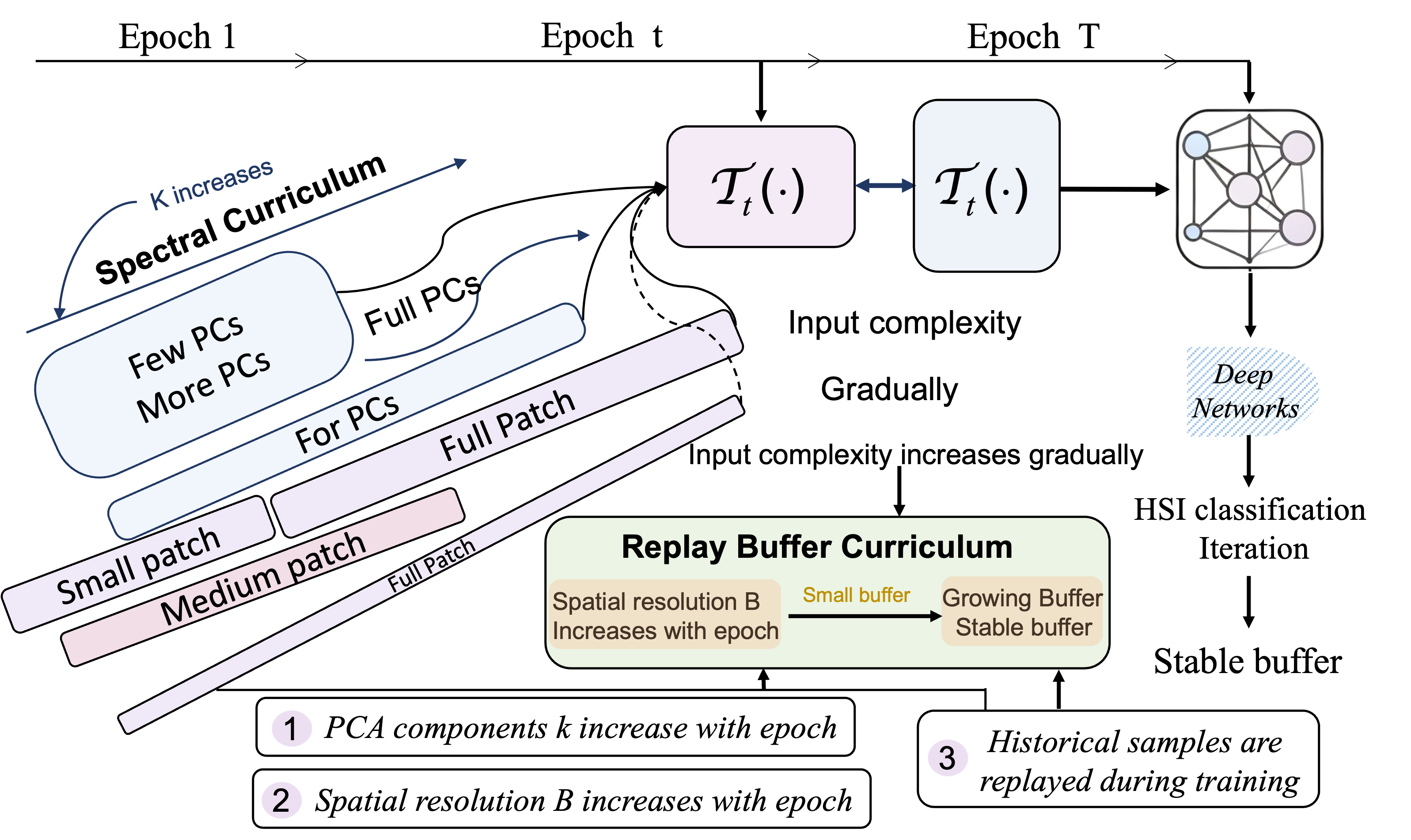}
  \caption{\textcolor{blue}{Overview of the dual curriculum learning framework for hyperspectral image classification and training process. Along the training epoch axis (1→t→T), the spectral curriculum increases retained PCA components k, and the spatial curriculum enlarges input patch resolution B. The epoch-adaptive transform \(\mathcal{T}_t(\cdot)\) controls input complexity. A replay buffer curriculum further reuses historical samples by expanding buffer capacity during training. Transformed samples are sent to deep networks for HSI classification, and three key curriculum rules are listed below the diagram.}}
\label{fig:trainprocess}
\end{figure}

SpectralTrain couples a spectral curriculum with a compute-balanced step allocation. At stage \(i\), we (i) select or search a spatial resolution \(B_i\) (monotonic non-decreasing over stages) and (ii) monotonically increase \(k_i\) (e.g., linear, piecewise-linear, or validation-guided). PCA is computed on the CPU (per image or per scene) and applied before GPU upload; resizing to \(B_i\) is then performed.

We also allow a short proxy fine-tuning of \(T_{\mathrm{ft}}\) epochs to select \(B_i\) from a candidate set \(\mathcal{B}_i\) under the fixed budget, extending the resolution-order search idea to hyperspectral training.

\begin{algorithm}[]
\caption{\textcolor{blue}{SpectralTrain}}
\label{alg:spectraltrain}
\color{blue}
\begin{algorithmic}[1]
\Require
Training set $\mathcal{D}_{\mathrm{tr}}$;
validation set $\mathcal{D}_{\mathrm{val}}$;
final spatial size $B_0$;
original spectral dimension $D$;
maximum retained PCA rank $K_{\max}\leq D$;
baseline epochs $T_0$;
budget ratio $\beta\in(0,1]$;
number of stages $N$;
proxy fine-tuning length $T_{\mathrm{ft}}$;
candidate spatial sets $\{\mathcal{B}_i\}_{i=1}^{N}$;
spectral schedule
$k_1\leq\cdots\leq k_N=K_{\max}$;
learning-rate scheduler $\alpha_{\mathrm{lr}}$.
\Ensure
Trained parameters $\Theta_N$ and stage-wise configuration
$\{(B_i,k_i)\}_{i=1}^{N}$.
\State $T\Leftarrow\beta T_0$.
\State Fit a PCA basis
$E_{K_{\max}}\in\mathbb{R}^{D\times K_{\max}}$
using only $\mathcal{D}_{\mathrm{tr}}$.
\State Initialize $\Theta_0$ with a fixed
$K_{\max}$-channel input operator.
\State $\Theta\Leftarrow\Theta_0$.
\For{$i\Leftarrow1$ to $N$}
    \Statex
    \State \textbf{Select the stage spatial size}
    \If{$|\mathcal{B}_i|>1$}
        \ForAll{$B\in\mathcal{B}_i$}
            \State
            $\Theta^{(B)}
            \Leftarrow
            \Call{Copy}{\Theta}$.
            \State
            $\mathcal{D}^{(B)}_i
            \Leftarrow
            \Call{ReduceAndResize}
            {
                \mathcal{D}_{\mathrm{tr}},
                E_{K_{\max}},
                k_i,
                B
            }$.
            \State
            $\Theta^{(B)}
            \Leftarrow
            \Call{FineTune}
            {
                \Theta^{(B)},
                \mathcal{D}^{(B)}_i,
                T_{\mathrm{ft}},
                \alpha_{\mathrm{lr}}
                \left(
                \frac{i-1}{N}
                :
                \frac{i-1}{N}
                +
                \frac{T_{\mathrm{ft}}}{T}
                \right)
            }$.
            \State
            $a_B
            \Leftarrow
            \Call{ValAcc}
            {
                \Theta^{(B)},
                \mathcal{D}_{\mathrm{val}},
                E_{K_{\max}},
                k_i,
                B
            }$.
        \EndFor
        \State
        $B_i
        \Leftarrow
        \displaystyle
        \arg\max_{B\in\mathcal{B}_i}a_B$.
    \Else
        \State
        $B_i\Leftarrow$
        the unique element of $\mathcal{B}_i$.
    \EndIf
    \Statex
    \State \textbf{Allocate the stage computation}
    \State
    $\mathrm{Steps}_i
    \Leftarrow
    \left\lfloor
    \dfrac{T}{N}
    \left(
    \dfrac{B_0}{B_i}
    \right)^2
    \right\rfloor$.
    \Statex
    \State \textbf{Train with the active spectral representation}
    \For{$s\Leftarrow1$ to $\mathrm{Steps}_i$}
        \State
        Sample a mini-batch
        $\{(X^{(b)},y^{(b)})\}_{b=1}^{m}$
        from $\mathcal{D}_{\mathrm{tr}}$.
        \For{$b\Leftarrow1$ to $m$}
            \State
            $Z^{(b)}_i
            \Leftarrow
            \Call{PCAProject}
            {
                X^{(b)},
                E_{K_{\max}},
                k_i
            }$.
            \Comment{materialize the leading $k_i$ PCs}
            \State
            $\widetilde{Z}^{(b)}_i
            \Leftarrow
            \Call{ResizeSpatial}
            {
                Z^{(b)}_i,
                B_i
            }$.
        \EndFor
        \State
        Transfer
        $\{(\widetilde{Z}^{(b)}_i,y^{(b)})\}_{b=1}^{m}$
        to the GPU.
        \State
        $\Theta
        \Leftarrow
        \Call{ActiveInputStep}
        {
            \Theta,
            \{(\widetilde{Z}^{(b)}_i,y^{(b)})\}_{b=1}^{m},
            k_i,
            K_{\max},
            \alpha_{\mathrm{lr}}
            \left(
            \frac{i-1}{N}
            :
            \frac{i}{N}
            \right)
        }$.
        \Comment{evaluate the first $k_i$ input slices}
    \EndFor
    \State $\Theta_i\Leftarrow\Theta$.
\EndFor
\State \Return
$\Theta_N$ and
$\{(B_i,k_i)\}_{i=1}^{N}$.
\end{algorithmic}
\end{algorithm}

\textcolor{blue}{In Algorithm~\ref{alg:spectraltrain},
$\mathrm{ActiveInputStep}$ evaluates only the first $k_i$ slices of the
fixed $K_{\max}$-channel input operator. The remaining input slices and
their optimizer states are retained without being updated during stage
$i$. All downstream layers preserve the same parameterization throughout
training. Increasing $k_i$ therefore activates additional spectral slices
without replacing the input layer, reinitializing the backbone, or
resetting the optimizer state.}

\subsection{Constraint Factors and Training Process}\label{subsec:process}

\paragraph{Resolution-controlled steps.}
Unlike fixed-size training, SpectralTrain adapts \(B_i\) per stage and balances compute using Eq.~\eqref{eq:steps}. Intuitively, smaller \(B_i\) yields more parameter updates early on (lower per-step cost), while larger \(B_i\) later preserves fine spatial structure when the model is ready to absorb it.

\paragraph{Spectral compression prior to upload.}
Before transferring data to GPU, we apply PCA to compress the spectral axis to \(k_i\) components, reducing I/O and VRAM pressure with ratio \(D/k_i\) in Eq.~\eqref{eq:cr}. PCA can be computed per image or per scene on CPU and cached; the projection is applied on the fly to each mini-batch.

\paragraph{Stage switching.}
Stages are synchronized with the scheduler \(\alpha_{\mathrm{lr}}\) over \(((i-1)/N,i/N]\). At the boundary, we (optionally) perform a short fine-tuning of \(T_{\mathrm{ft}}\) epochs at \((B_i,k_i)\) to smooth the transition before moving to stage \(i{+}1\).

{\color{blue}

\subsection{Theoretical Analysis}
\label{subsec:theory}

Let $\widetilde{x}=x-\mu\in\mathbb{R}^{D}$ denote a centered spectral
vector sampled from the training distribution. Its spectral covariance is
\begin{equation}
C
=
\mathbb{E}\!\left[\widetilde{x}\widetilde{x}^{\top}\right]
=
E\Lambda E^{\top},
\qquad
\Lambda
=
\operatorname{diag}(\lambda_1,\ldots,\lambda_D),
\label{eq:theory_covariance}
\end{equation}
where
\(
\lambda_1\geq\lambda_2\geq\cdots\geq\lambda_D\geq0
\)
and
\(
E=[e_1,\ldots,e_D]
\)
is orthogonal. For a retained rank \(k\), define
\begin{equation}
E_k=[e_1,\ldots,e_k],
\qquad
P_k=E_kE_k^{\top}.
\label{eq:theory_projector}
\end{equation}
For comparison across spectral ranks, we also define the zero-padded PCA
coordinate representation
\begin{equation}
z_k(\widetilde{x})
=
\begin{bmatrix}
E_k^{\top}\widetilde{x}\\
\mathbf{0}_{D-k}
\end{bmatrix}
\in\mathbb{R}^{D}.
\label{eq:zero_padded_pca}
\end{equation}
This common coordinate space is used only for the following analysis. The
implementation stores and transfers the active \(k\) coordinates.

\begin{theorem}
\label{thm:nested_pca}
For any \(1\leq k\leq k'\leq D\), the PCA projectors satisfy
\begin{equation}
P_kP_{k'}=P_k.
\label{eq:nested_projector}
\end{equation}
The expected spectral reconstruction error is
\begin{equation}
\mathbb{E}
\left[
\left\|
\widetilde{x}-P_k\widetilde{x}
\right\|_2^2
\right]
=
\sum_{j=k+1}^{D}\lambda_j.
\label{eq:pca_expected_tail}
\end{equation}
Consequently,
\begin{equation}
\mathbb{E}
\left[
\left\|
\widetilde{x}-P_{k'}\widetilde{x}
\right\|_2^2
\right]
\leq
\mathbb{E}
\left[
\left\|
\widetilde{x}-P_k\widetilde{x}
\right\|_2^2
\right].
\label{eq:pca_monotone_tail}
\end{equation}
Among all rank-\(k\) orthogonal projectors, \(P_k\) minimizes the expected
squared reconstruction error.
\end{theorem}

\begin{proof}
Since the columns of \(E\) form an orthonormal basis and the first \(k\)
columns of \(E_k\) are contained in \(E_{k'}\), we have
\begin{equation}
E_k^{\top}E_{k'}
=
\begin{bmatrix}
I_k & 0
\end{bmatrix}.
\end{equation}
It follows that
\begin{align}
P_kP_{k'}
&=
E_kE_k^{\top}E_{k'}E_{k'}^{\top} \\
&=
E_k
\begin{bmatrix}
I_k & 0
\end{bmatrix}
E_{k'}^{\top} \\
&=
E_kE_k^{\top}
=
P_k.
\end{align}

Using the eigendecomposition in
Eq.~\eqref{eq:theory_covariance}, the expected error associated with an
orthogonal projector \(Q\) is
\begin{align}
\mathbb{E}
\left[
\left\|
\widetilde{x}-Q\widetilde{x}
\right\|_2^2
\right]
&=
\operatorname{tr}
\left[
(I-Q)C(I-Q)^{\top}
\right].
\end{align}
For \(Q=P_k\), idempotence and symmetry give
\begin{align}
\mathbb{E}
\left[
\left\|
\widetilde{x}-P_k\widetilde{x}
\right\|_2^2
\right]
&=
\operatorname{tr}\!\left[(I-P_k)C\right] \\
&=
\operatorname{tr}(C)
-
\operatorname{tr}(P_kC) \\
&=
\sum_{j=1}^{D}\lambda_j
-
\sum_{j=1}^{k}\lambda_j \\
&=
\sum_{j=k+1}^{D}\lambda_j.
\end{align}
The tail sum decreases as \(k\) increases, which proves
Eq.~\eqref{eq:pca_monotone_tail}.

For any rank-\(k\) orthogonal projector \(Q\),
\begin{equation}
\mathbb{E}
\left[
\left\|
\widetilde{x}-Q\widetilde{x}
\right\|_2^2
\right]
=
\operatorname{tr}(C)-\operatorname{tr}(QC).
\end{equation}
The Ky Fan maximum principle gives
\begin{equation}
\operatorname{tr}(QC)
\leq
\sum_{j=1}^{k}\lambda_j.
\end{equation}
Equality is attained by \(Q=P_k\). Therefore, \(P_k\) minimizes the
expected squared reconstruction error among all rank-\(k\) orthogonal
projectors.
\end{proof}

\begin{remark}
The theorem is independent of the classifier and optimizer. It establishes
that the spectral curriculum forms a nested sequence of subspaces and that
every increase in \(k_i\) preserves the previously exposed PCA directions.
The discarded spectral energy decreases from
\(
\sum_{j=k_i+1}^{D}\lambda_j
\)
to zero when the final stage uses \(k_N=D\). If the implementation adopts
a maximum retained rank \(K_{\max}<D\), the final residual energy equals
\(
\sum_{j=K_{\max}+1}^{D}\lambda_j
\).
\end{remark}

\begin{theorem}
\label{thm:input_cost}
Consider a mini-batch of \(m\) hyperspectral inputs at stage \(i\). The
stage uses \(k_i\) active PCA components and a spatial size
\(B_i\times B_i\). When only active components are materialized, the input
volume is
\begin{equation}
V_i=mB_i^2k_i.
\label{eq:stage_volume}
\end{equation}
Relative to the final configuration \((D,B_0)\), the input-volume ratio is
\begin{equation}
\frac{V_i}{V_0}
=
\frac{k_i}{D}
\left(
\frac{B_i}{B_0}
\right)^2.
\label{eq:stage_volume_ratio}
\end{equation}
For a convolutional stem with fixed kernel size and output width, or a
linear patch embedding with fixed output width, the input-dependent
multiply--accumulate count follows the same ratio.
\end{theorem}

\begin{proof}
Each sample contains \(B_i^2\) spatial positions and \(k_i\) active
spectral coefficients. A mini-batch therefore contains
\begin{equation}
mB_i^2k_i
\end{equation}
scalar values. The final full-complexity configuration contains
\begin{equation}
mB_0^2D
\end{equation}
values. Division gives Eq.~\eqref{eq:stage_volume_ratio}.

For a convolutional stem with kernel size \(q\times q\) and \(c\) output
channels, the leading input-dependent operation count is proportional to
\begin{equation}
mB_i^2q^2k_ic.
\end{equation}
The quantities \(m\), \(q\), and \(c\) remain fixed across stages, so the
relative count is again
\begin{equation}
\frac{k_i}{D}
\left(
\frac{B_i}{B_0}
\right)^2.
\end{equation}
A linear patch embedding has the same dependence on the number of spatial
positions and active input components.
\end{proof}

\begin{remark}
Equation~\eqref{eq:stage_volume_ratio} gives an exact count for input
elements and an exact ratio for the input-dependent part of the stated
operators. Total wall-clock time also includes data loading, fixed backbone
operations, memory access, kernel launch, and hardware-dependent overhead.
The measured acceleration therefore remains an empirical systems result,
while the theorem identifies the component of the cost that SpectralTrain
directly controls.
\end{remark}

\begin{theorem}
\label{thm:objective_discrepancy}
Let \(f_{\Theta}\) be a network with parameters \(\Theta\), and define the
rank-\(k\) objective
\begin{equation}
\mathcal{L}_k(\Theta)
=
\mathbb{E}_{(\widetilde{x},y)}
\left[
\ell
\left(
f_{\Theta}(z_k(\widetilde{x})),
y
\right)
\right].
\label{eq:rank_k_objective}
\end{equation}
Assume that \(f_{\Theta}\) is locally \(L_f(\Theta)\)-Lipschitz with
respect to its input over the considered data domain, and that
\(\ell(\cdot,y)\) is \(L_{\ell}\)-Lipschitz with respect to the network
output. Then
\begin{equation}
\left|
\mathcal{L}_k(\Theta)
-
\mathcal{L}_D(\Theta)
\right|
\leq
L_{\ell}L_f(\Theta)
\left(
\sum_{j=k+1}^{D}\lambda_j
\right)^{1/2}.
\label{eq:objective_discrepancy}
\end{equation}

Assume further that
\begin{equation}
g_{\Theta}(z,y)
=
\nabla_{\Theta}
\ell
\left(
f_{\Theta}(z),y
\right)
\end{equation}
is locally \(G_{\Theta}\)-Lipschitz with respect to \(z\). Then
\begin{equation}
\left\|
\nabla_{\Theta}\mathcal{L}_k(\Theta)
-
\nabla_{\Theta}\mathcal{L}_D(\Theta)
\right\|
\leq
G_{\Theta}
\left(
\sum_{j=k+1}^{D}\lambda_j
\right)^{1/2}.
\label{eq:gradient_discrepancy}
\end{equation}
\end{theorem}

\begin{proof}
The Lipschitz assumptions yield
\begin{align}
&
\left|
\mathcal{L}_k(\Theta)
-
\mathcal{L}_D(\Theta)
\right|
\nonumber\\
&\leq
\mathbb{E}
\left[
\left|
\ell
\left(
f_{\Theta}(z_k),y
\right)
-
\ell
\left(
f_{\Theta}(z_D),y
\right)
\right|
\right]
\nonumber\\
&\leq
L_{\ell}
\mathbb{E}
\left[
\left\|
f_{\Theta}(z_k)
-
f_{\Theta}(z_D)
\right\|
\right]
\nonumber\\
&\leq
L_{\ell}L_f(\Theta)
\mathbb{E}
\left[
\left\|
z_k-z_D
\right\|
\right].
\end{align}
The Cauchy--Schwarz inequality gives
\begin{equation}
\mathbb{E}\|z_k-z_D\|
\leq
\left(
\mathbb{E}\|z_k-z_D\|^2
\right)^{1/2}.
\end{equation}
The PCA coordinates are orthogonal, so
\begin{align}
\mathbb{E}
\left[
\left\|
z_k-z_D
\right\|_2^2
\right]
&=
\mathbb{E}
\left[
\sum_{j=k+1}^{D}
(e_j^{\top}\widetilde{x})^2
\right] \\
&=
\sum_{j=k+1}^{D}\lambda_j.
\end{align}
Substitution proves Eq.~\eqref{eq:objective_discrepancy}.

For the gradient, Jensen's inequality and the assumed Lipschitz property
give
\begin{align}
&
\left\|
\nabla_{\Theta}\mathcal{L}_k(\Theta)
-
\nabla_{\Theta}\mathcal{L}_D(\Theta)
\right\|
\nonumber\\
&=
\left\|
\mathbb{E}
\left[
g_{\Theta}(z_k,y)
-
g_{\Theta}(z_D,y)
\right]
\right\|
\nonumber\\
&\leq
\mathbb{E}
\left[
\left\|
g_{\Theta}(z_k,y)
-
g_{\Theta}(z_D,y)
\right\|
\right]
\nonumber\\
&\leq
G_{\Theta}
\mathbb{E}
\left[
\left\|
z_k-z_D
\right\|
\right]
\nonumber\\
&\leq
G_{\Theta}
\left(
\sum_{j=k+1}^{D}\lambda_j
\right)^{1/2}.
\end{align}
This proves Eq.~\eqref{eq:gradient_discrepancy}.
\end{proof}

\begin{remark}
This theorem does not require convexity of the network objective. It applies
to a differentiable deep network at a fixed parameter state whenever the
stated local Lipschitz constants are finite. The bounds show that the
stage-wise objective and parameter gradient approach their full-rank
counterparts as \(k_i\) increases. They provide no claim that the resulting
non-convex optimization trajectory reaches a global minimum or preserves
classification accuracy for every architecture.

The spatial curriculum can be incorporated through an analysis
reconstruction operator. Let
\(\mathcal{S}_{B_i}\) denote spatial reduction and
\(\mathcal{R}_{B_i}\) denote reconstruction to the final spatial grid. If
\begin{equation}
\delta_i
=
\left(
\mathbb{E}
\left[
\left\|
P_{k_i}\widetilde{x}
-
\mathcal{R}_{B_i}
\mathcal{S}_{B_i}
P_{k_i}\widetilde{x}
\right\|_2^2
\right]
\right)^{1/2},
\end{equation}
then the total input discrepancy is bounded by
\begin{equation}
\left(
\sum_{j=k_i+1}^{D}\lambda_j
\right)^{1/2}
+
\delta_i.
\label{eq:joint_discrepancy}
\end{equation}
Equations~\eqref{eq:objective_discrepancy} and
\eqref{eq:gradient_discrepancy} extend by replacing the spectral-tail term
with Eq.~\eqref{eq:joint_discrepancy}.
\end{remark}

\begin{theorem}
\label{thm:surrogate_continuation}
Let
\(
F(\Theta)=\mathcal{L}_D(\Theta)
\)
be an \(L\)-smooth and \(\mu\)-strongly convex surrogate objective with
minimizer \(\Theta^{\star}\). Consider the stage-\(i\) update
\begin{equation}
\Theta_{t+1}
=
\Theta_t
-
\eta
\nabla_{\Theta}\mathcal{L}_{k_i}(\Theta_t),
\label{eq:stage_update}
\end{equation}
and assume
\begin{equation}
\left\|
\nabla_{\Theta}\mathcal{L}_{k_i}(\Theta)
-
\nabla F(\Theta)
\right\|
\leq
\varepsilon_i
\label{eq:stage_gradient_error}
\end{equation}
over the considered stage. Set
\begin{equation}
\eta=\frac{2}{L+\mu},
\qquad
q=\frac{L-\mu}{L+\mu}.
\label{eq:surrogate_step}
\end{equation}
After \(s\) stage updates,
\begin{equation}
\left\|
\Theta_{t+s}
-
\Theta^{\star}
\right\|
\leq
q^s
\left\|
\Theta_t-\Theta^{\star}
\right\|
+
\frac{\varepsilon_i}{\mu}
\left(
1-q^s
\right).
\label{eq:surrogate_bound}
\end{equation}
Under Theorem~\ref{thm:objective_discrepancy}, one may use
\begin{equation}
\varepsilon_i
=
G_{\Theta}
\left(
\sum_{j=k_i+1}^{D}\lambda_j
\right)^{1/2}
\label{eq:surrogate_error_radius}
\end{equation}
when the corresponding local gradient Lipschitz condition holds.
\end{theorem}

\begin{proof}
Define
\begin{equation}
e_t
=
\nabla_{\Theta}\mathcal{L}_{k_i}(\Theta_t)
-
\nabla F(\Theta_t).
\end{equation}
Equation~\eqref{eq:stage_update} becomes
\begin{equation}
\Theta_{t+1}
=
\Theta_t
-
\eta
\left[
\nabla F(\Theta_t)+e_t
\right],
\qquad
\|e_t\|\leq\varepsilon_i.
\end{equation}
For an \(L\)-smooth and \(\mu\)-strongly convex objective, gradient descent
with the step size in Eq.~\eqref{eq:surrogate_step} satisfies
\begin{equation}
\left\|
\Theta_t
-
\eta\nabla F(\Theta_t)
-
\Theta^{\star}
\right\|
\leq
q
\left\|
\Theta_t-\Theta^{\star}
\right\|.
\end{equation}
The triangle inequality therefore gives
\begin{align}
\left\|
\Theta_{t+1}-\Theta^{\star}
\right\|
&\leq
q
\left\|
\Theta_t-\Theta^{\star}
\right\|
+
\eta\varepsilon_i.
\label{eq:surrogate_recursion}
\end{align}
Applying Eq.~\eqref{eq:surrogate_recursion} recursively for \(s\) updates
yields
\begin{align}
\left\|
\Theta_{t+s}-\Theta^{\star}
\right\|
&\leq
q^s
\left\|
\Theta_t-\Theta^{\star}
\right\|
+
\eta\varepsilon_i
\sum_{r=0}^{s-1}q^r \\
&=
q^s
\left\|
\Theta_t-\Theta^{\star}
\right\|
+
\eta\varepsilon_i
\frac{1-q^s}{1-q}.
\end{align}
Since
\begin{equation}
1-q
=
\frac{2\mu}{L+\mu}
=
\eta\mu,
\end{equation}
we obtain
\begin{equation}
\eta\varepsilon_i
\frac{1-q^s}{1-q}
=
\frac{\varepsilon_i}{\mu}
\left(
1-q^s
\right),
\end{equation}
which proves Eq.~\eqref{eq:surrogate_bound}.
\end{proof}

\begin{remark}
The theorem provides an explanatory continuation model under a smooth
strongly convex surrogate. It shows that a low-rank stage can approach a
neighborhood of the final-rank solution and that the neighborhood radius
decreases as the discarded spectral energy decreases. Deep convolutional
and transformer objectives generally lack global strong convexity.
Accordingly, Theorem~\ref{thm:surrogate_continuation} supplies a
mechanism-level interpretation. The cross-backbone and cross-dataset
experiments provide the evidence for the deep non-convex setting.
\end{remark}

}

\subsection{Computational Constraint Order Search}\label{subsec:ordersearch}

We extend the resolution order-search idea to hyperspectral training by selecting \(B_i\) from \(\mathcal{B}_i\) with a short validation-guided proxy, while \(k_i\) follows a monotonic schedule (e.g., linear from \(k_1\) to \(D\)). At stage \(i\), we choose
\begin{equation}
\widehat{B}_i
=
\arg\max_{B\in \mathcal{B}_i}\ 
\mathrm{ValAcc}\!\left(
\mathrm{FineTune}\!\left(
\Theta_{i-1},\,
\mathrm{ReduceAndResize}(\mathcal{D},k_{i},B),\,
T_{\mathrm{ft}},\,
\alpha_{\mathrm{lr}}
\right)
\right),
\label{eq:argmaxB}
\end{equation}
then train for \(\mathrm{Steps}_i\) steps computed by Eq.~\eqref{eq:steps}. The above design decouples \emph{what} the model sees (spectral/spatial complexity) from \emph{how much} compute it receives (steps), enabling progressive optimization under a fixed budget without modifying architectures or losses.

\section{Results}\label{sec:results}

\subsection{Experimental Setup}

The evaluation of SpectralTrain is conducted under controlled and comparable
conditions. Three datasets with different spatial sizes, spectral dimensions,
and scene characteristics are considered. Indian Pines contains a
$145\times145$ scene with 200 spectral bands, Salinas-A contains an
$86\times83$ scene with 204 bands, and CloudPatch-7 contains
$50\times50$ hyperspectral patches with 462 bands.
\textcolor{blue}{CloudPatch-7 comprises 380 annotated patches extracted
from 28 parent scenes and covers wavelengths from 400 to 1000\,nm.}
Training uses PyTorch~2.1.2, TorchVision~0.16.2, and
TorchAudio~2.1.2 with CUDA~12.1 and Python~3.10. All experiments are
conducted on an Intel Core i7-13700 processor with 16 threads, 32\,GB RAM,
an NVIDIA RTX~4090D GPU with 24\,GB memory, and a 50\,GB NVMe SSD
for preprocessing and caching.

The evaluated architectures include classical convolutional neural networks
(CNNs), including ResNet-34 \cite{24} and ConvNeXt-T \cite{25}, hybrid
CNN--Transformer models, including MetaFormer \cite{26} and ADGAN
\cite{27}, and spectral-oriented models, including 3D-ConvSST \cite{28}
and SQS \cite{29}. DSFormer \cite{30} is included as a recent
hyperspectral classification reference. A conventional full-complexity
training protocol \cite{31} provides the baseline for the efficiency
comparison. Overall Accuracy (OA), Average Accuracy (AA), and Cohen's
Kappa are used as the evaluation metrics.

\textcolor{blue}{For a fair comparison, conventional training and
SpectralTrain use the same data split, backbone, maximum retained PCA rank,
final spatial size, batch size, loss function, optimizer, regularization
parameters, learning-rate schedule, training seed, and total equivalent
training budget. The conventional protocol uses the maximum retained PCA
rank and final spatial size throughout training. SpectralTrain changes only
the active PCA rank and spatial input size according to the prescribed
curriculum. All timing measurements are obtained under the same hardware
and software environment.}

\textcolor{blue}{Following the evaluation protocol in \cite{31}, Indian
Pines and Salinas-A use center-labeled spatial patches extracted with unit
stride. Their final patch sizes are $22\times22$ and $18\times18$,
respectively. The extracted patch sets are divided into 80\% training
samples and 20\% test samples using random seed 42. This within-scene
protocol is retained to ensure direct comparability with \cite{31}. The
principal component analysis (PCA) basis is fitted using the training
subset and is then applied unchanged to the test subset.}

\textcolor{blue}{CloudPatch-7 is divided at the parent-scene level because
multiple hyperspectral patches originate from the same full-scene capture.
The 28 parent scenes are assigned to 20 training scenes, 3 validation
scenes, and 5 test scenes under a fixed grouped split. All patches from one
parent scene remain in the same subset. The validation subset supports
curriculum configuration, and the test subset is used only for final
evaluation. The PCA basis is fitted using the training scenes and is
applied unchanged to the validation and test scenes. This grouping prevents
parent-scene content from being shared across the three subsets.}

\textcolor{blue}{SpectralTrain uses a three-stage spectral--spatial
curriculum with a maximum retained PCA rank of $K_{\max}=30$. The spectral
schedule is
\[
(k_1,k_2,k_3)=(10,20,30),
\]
and the spatial schedule is
\[
\left(
\frac{B_1}{B_0},
\frac{B_2}{B_0},
\frac{B_3}{B_0}
\right)
=
(0.45,0.70,1.00).
\]
The corresponding spatial sizes are $(10,15,22)$ for Indian Pines,
$(8,13,18)$ for Salinas-A, and $(14,21,30)$ for CloudPatch-7.
Spatial sizes obtained from the shared ratios are rounded to the nearest
integer. The dataset-specific configurations are summarized in
Table~\ref{tab:dataset_config}.}

\begin{table}[t]
\centering
\caption{\textcolor{blue}{Dataset-specific configurations of SpectralTrain.
$B_0$ denotes the final patch size, and $(B_1,B_2,B_3)$ gives the
spatial sizes used in the three curriculum stages.}}
\label{tab:dataset_config}
\scriptsize
\setlength{\tabcolsep}{5pt}
\begin{tabular*}{\textwidth}{
@{\extracolsep{\fill}}lccccc}
\toprule
Dataset &
Data unit &
Bands &
Classes &
$B_0$ &
$(B_1,B_2,B_3)$ \\
\midrule
Indian Pines &
$145\times145$ scene &
200 &
16 &
22 &
$(10,15,22)$ \\

Salinas-A &
$86\times83$ scene &
204 &
6 &
18 &
$(8,13,18)$ \\

CloudPatch-7 &
380 patches from 28 scenes &
462 &
7 &
30 &
$(14,21,30)$ \\
\bottomrule
\end{tabular*}
\end{table}

\textcolor{blue}{The PCA components are ordered by decreasing explained
variance. Each backbone retains an input parameterization for
$K_{\max}$ components. At stage $i$, only the leading $k_i$ PCA
coefficients and their corresponding input-weight slices are evaluated.
Later stages progressively activate additional component slices while
preserving the existing model parameters and optimizer states. This
implementation maintains a continuous training trajectory without replacing
the input layer or reinitializing the network. Convolutional backbones
directly process the stage-dependent spatial sizes. Position-dependent
quantities in transformer-based backbones are resized when the spatial
input changes.}

\textcolor{blue}{Both conventional training and SpectralTrain are assigned
$T_0=T=300$ equivalent full-complexity epochs, which gives $\beta=1$.
The training process is divided equally into $N=3$ stages. The three
equivalent-budget intervals are 1--100, 101--200, and 201--300.
Within each stage, the number of mini-batch updates follows
Eq.~\eqref{eq:steps}. Smaller spatial inputs therefore permit more updates
under the same assigned equivalent computation. The learning-rate schedule
is indexed by the consumed equivalent budget, which preserves the same
global learning-rate trajectory for conventional training and
SpectralTrain. The shared stage-wise schedule is summarized in
Table~\ref{tab:curriculum_config}.}

\begin{table}[t]
\centering
\caption{\textcolor{blue}{Shared stage-wise configuration of SpectralTrain.
Both conventional training and SpectralTrain use 300 equivalent
full-complexity epochs, with $N=3$ and $\beta=1$. The proxy length
$T_{\mathrm{ft}}$ is used only during the one-time curriculum
configuration procedure.}}
\label{tab:curriculum_config}
\small
\setlength{\tabcolsep}{5.5pt}
\begin{tabular*}{\textwidth}{
@{\extracolsep{\fill}}ccccc}
\toprule
Stage &
Equivalent-budget interval &
Active PCs $k_i$ &
$B_i/B_0$ &
$T_{\mathrm{ft}}$ \\
\midrule
1 & 1--100   & 10 & 0.45 & 10 \\
2 & 101--200 & 20 & 0.70 & 10 \\
3 & 201--300 & 30 & 1.00 & -- \\
\bottomrule
\end{tabular*}
\end{table}

\textcolor{blue}{The reported schedule is obtained through a one-time
validation-guided configuration procedure and is then applied unchanged to
all evaluated backbones and datasets. Candidate spatial configurations in
the first two stages are evaluated using
$T_{\mathrm{ft}}=10$ proxy fine-tuning epochs. The proxy procedure
operates on copied checkpoints, while the main curriculum retains the
corresponding stage checkpoint. The one-time configuration cost is outside
the repeated per-backbone training comparison and is amortized across all
evaluated models. The final stage already uses $(K_{\max},B_0)$ and
requires no proxy evaluation. The replay factor is fixed at one in all
reported experiments.}

All experiments use a batch size of 16. The optimizer is AdamW with an
initial learning rate of $4\times10^{-3}$,
$\epsilon=1\times10^{-8}$, and a weight decay of 0.05. The cosine
learning-rate schedule uses 20 warm-up epochs and a minimum learning rate
of $1\times10^{-6}$. Label smoothing is set to 0.1. Layer decay is set
to 1.0, drop path and head dropout are set to zero, and the layer-scale
initialization value is $1\times10^{-6}$. The gradient update frequency
is one, and the training seed is fixed at zero.
\textcolor{blue}{The 20-epoch warm-up is applied identically to
conventional training and SpectralTrain. It controls the global
learning-rate trajectory and remains independent of the
$T_{\mathrm{ft}}=10$ proxy configuration procedure.}
Data splits, optimization settings, loss functions, and learning-rate
schedules remain unchanged within each controlled comparison.

\subsection{Ablative Study}
\begin{figure}[t]
\centering
\includegraphics[width=\linewidth]{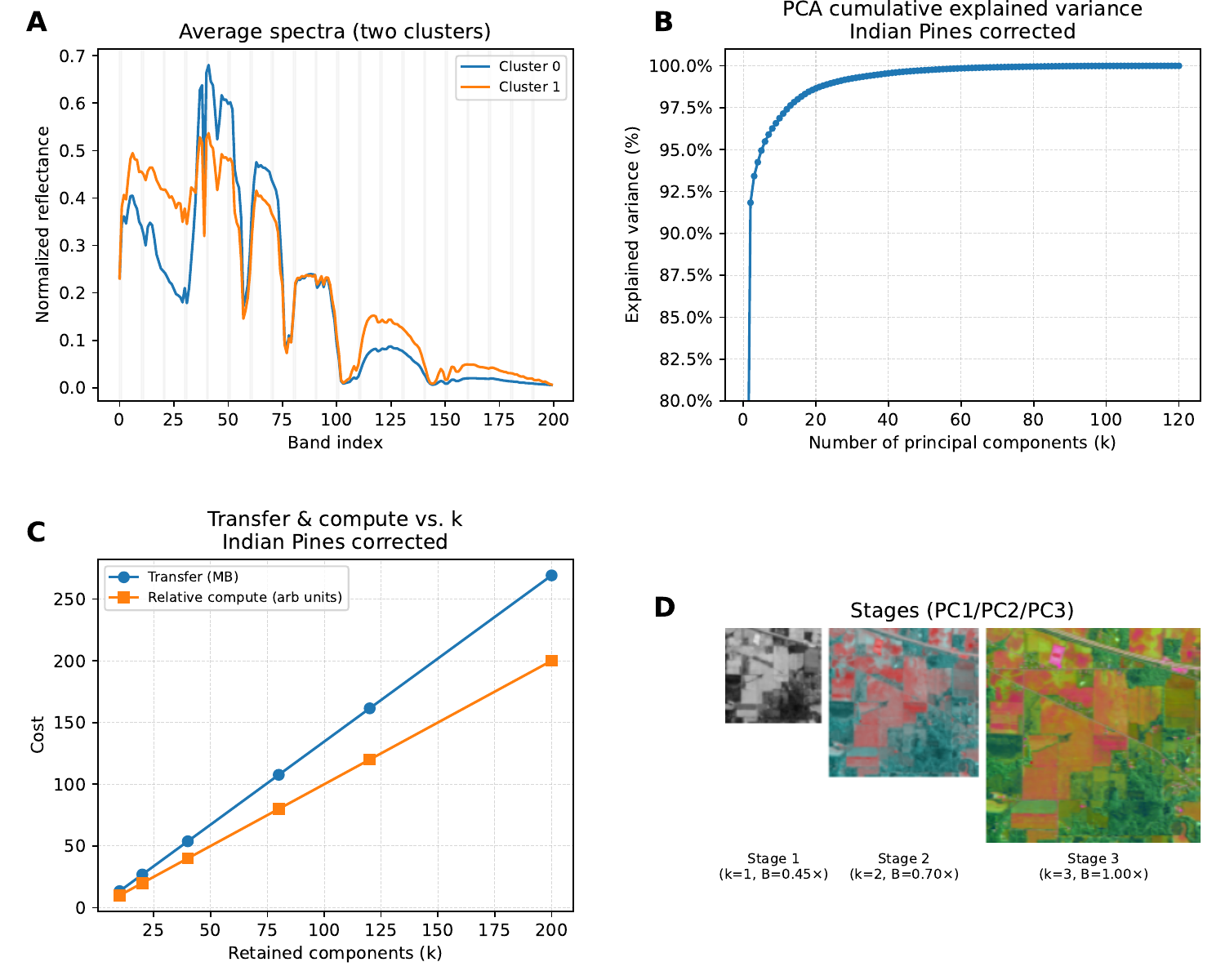}
\caption{Motivation example for a spectral curriculum on Indian Pines. 
(A) Average spectra of two clusters show band-localized discriminative cues, so uniform band dropping risks removing signals. 
(B) PCA cumulative explained variance illustrates that a small number of components preserve most energy, motivating an information-preserving low-cost start; PCA is used here as a representative linear compressor. 
(C) Per-epoch transfer and compute scale with the number of retained components $k$, so beginning with small $k$ saves early cost. 
(D) Curriculum stages progressively increase both spectral components and spatial size (PC1/PC2/PC3 composites). 
Panels illustrate one instantiation with PCA; Table~\ref{tab:dr} shows that alternative reducers (UMAP/ICA) produce comparable behavior under the same staged schedule.}
\label{fig:motivation}
\end{figure}

Figure~\ref{fig:motivation} summarizes the empirical motivation for a spectral curriculum. The two cluster-mean spectra from Indian Pines in panel~A show that discriminative information concentrates in narrow, non-uniform wavelength intervals; removing bands uniformly at the beginning can erase these cues at random. Panel~B indicates that a small number of principal components captures most spectral energy on the same cube, which permits an information-preserving start. Panel~C shows that host-to-device transfer and per-epoch compute grow approximately linearly with the number of retained components $k$, so using a small $k$ early saves cost in the regime where gradients are still coarse. Panel~D visualizes staged inputs that progressively restore spectral and spatial complexity. In combination, these diagnostics support the decision to organize training along spectra rather than fixing full cubes from the first epoch, which follows the spirit of resolution curricula in computer vision \cite{8} while addressing hyperspectral specifics. 

All ablation experiments were conducted on the CloudPatch-7 Dataset. A first question is whether additional spectral filtering remains beneficial once information-preserving compression is applied. Table~\ref{tab:filters} keeps schedules fixed and appends Savitzky--Golay \cite{32}, Gaussian smoothing \cite{33}, or total-variation \cite{34} denoising on top of PCA-based compression. The figures indicate no gains in OA/AA or Kappa and, for Gaussian smoothing, higher epoch time. After PCA concentrates low-rank structure, hand-crafted smoothing becomes redundant while adding cost.

\begin{table}[t]
\caption{Performance with different filters added. Time denotes seconds per epoch.}
\label{tab:filters}
\centering
\small
\begin{tabular*}{\textwidth}{@{\extracolsep{\fill}}lcccc}
\toprule
Settings & OA (\%) & AA (\%) & Kappa & Time (s/epoch) \\
\midrule
Ours (PCA curriculum) & 99.02 & 99.04 & 0.99 & 8.00 \\
+ Savitzky--Golay     & 98.82 & 98.87 & 0.99 & 8.00 \\
+ Gaussian            & 98.80 & 98.84 & 0.99 & 11.00 \\
+ TV Denoising        & 98.87 & 98.91 & 0.99 & 8.00 \\
\bottomrule
\end{tabular*}
\end{table}

The schedule itself is then compared against a conventional loop \cite{31} across heterogeneous backbones while holding splits, optimizer, loss, and learning-rate schedule constant. Table~\ref{tab:traincompare} shows that SpectralTrain reduces seconds per epoch by about $2$--$7\times$ with comparable accuracy. For 3D-ConvSST, time drops from 28\,s to 4\,s while Kappa remains 0.99; for RCNN, time decreases from 38\,s to 6\,s with OA/AA near 99.44\%/99.42\%. This points to early-epoch staging along the spectral axis as an effective lever for efficiency that does not rely on architecture-specific tuning.

\begin{table}
\setlength{\tabcolsep}{0.5pt}
\caption{\textcolor{blue}{Comparison of conventional training and SpectralTrain using PCA downsampling. All models share the same splits and hyperparameters within each column group. Times measured on the same hardware; seeds fixed unless stated.}}
\label{tab:traincompare}
\centering
\scriptsize
\begin{tabular}{lcccccccc}
\toprule
\multirow{2}{*}{Model} &
\multicolumn{4}{c}{Current SOTA Training \cite{31}} &
\multicolumn{4}{c}{SpectralTrain} \\
\cmidrule(lr){2-5}\cmidrule(lr){6-9}
& Time (s/epoch) & OA (\%) & AA (\%) & Kappa
& Time (s/epoch) & OA (\%) & AA (\%) & Kappa \\
\midrule
ResNet-34 & 113 & 98.55 & 98.59 & 0.98 & 66 & 98.20 & 98.05 & 0.98 \\\\
ADGAN & 664 & 98.96 & 98.91 & 0.98 & 298 & 98.28 & 98.10 & 0.98 \\\\
MetaFormer & 77 & 98.93 & 98.98 & 0.99 & 17 & 98.95 & 98.94 & 0.99 \\\\
ConvNeXt-T & 269 & 99.06 & 99.14 & 0.99 & 94 & 99.15 & 99.18 & 0.99 \\\\
SQS & 356 & 99.30 & 99.30 & 0.99 & 254 & 98.17 & 98.11 & 0.98 \\\\
3D-ConvSST & 28 & 99.84 & 99.76 & 0.99 & 4 & 99.02 & 99.04 & 0.99 \\\\
RCNN & 38 & 99.75 & 99.76 & 0.99 & 6 & 99.44 & 99.42 & 0.99 \\\\
\bottomrule
\end{tabular}
\end{table}

Sensitivity to the compression operator was evaluated by replacing PCA with UMAP or ICA while keeping the curriculum stage boundaries, the learning–rate schedule, and the per-stage compute budgets fixed. As shown in Table~\ref{tab:dr}, all three reducers achieve comparable OA/AA at the same time per epoch, indicating that the staged restoration of spectral complexity—rather than the particular manifold mapping—drives the efficiency gains. When the retained dimensionality $k$ is very small, non-linear reducers can yield slightly larger validation margins for some classes; as $k$ increases and the full spectrum is restored, these differences quickly diminish, consistent with using the reducer only for a low-cost warm-up. Overall, SpectralTrain does not rely on the linearity of PCA and remains effective with any information-preserving compressor that is approximately nested across $k$ (i.e., the representation at $k$ is largely contained in that at $k'>k$).

\begin{table}[t]
\caption{Performance with different dimensionality reduction methods.}
\label{tab:dr}
\centering
\small
\begin{tabular*}{\textwidth}{@{\extracolsep{\fill}}lcccc}
\toprule
Dimensionality Reduction Method & OA (\%) & AA (\%) & Kappa & Time (s/epoch) \\
\midrule
PCA  & 99.15 & 99.18 & 0.99 & 8.00 \\
UMAP & 99.36 & 99.39 & 0.99 & 8.00 \\
ICA  & 98.89 & 98.86 & 0.99 & 8.00 \\
\bottomrule
\end{tabular*}
\end{table}

Loss definitions are compared next \cite{35,36,37}. Table~\ref{tab:loss} indicates that cross entropy, multi-label soft margin, and binary cross entropy with logits yield similar OA/AA and Kappa at the same epoch time. In standard practice, distinct losses can change margin behavior, calibration, and robustness, thereby altering performance; under the staged spectral schedule, these differences are attenuated, consistent with a smoother early optimization landscape.

\begin{table}[t]
\caption{Performance and efficiency of different loss functions.}
\label{tab:loss}
\centering
\small
\begin{tabular*}{\textwidth}{@{\extracolsep{\fill}}lcccc}
\toprule
Loss Function & OA (\%) & AA (\%) & Kappa & Time (s/epoch) \\
\midrule
Cross Entropy Loss               & 98.96 & 99.01 & 0.99 & 8.00 \\
Multi-Label Soft Margin Loss     & 98.96 & 98.97 & 0.99 & 8.00 \\
Binary Cross Entropy with Logits & 99.02 & 99.04 & 0.99 & 8.00 \\
\bottomrule
\end{tabular*}
\end{table}

Optimizer choice is evaluated in Table~\ref{tab:optim}. In conventional settings, AdamW \cite{38}, AdamMini \cite{39}, and SGD \cite{40} often exhibit different convergence speed and final accuracy because of distinct momentum and regularization properties. Under the spectral curriculum, all three converge effectively with small spread, indicating reduced sensitivity once early spectral complexity is aligned with learning progress.

\begin{table}[t]
\caption{Performance with different optimizers.}
\label{tab:optim}
\centering
\small
\begin{tabular*}{\textwidth}{@{\extracolsep{\fill}}lcccc}
\toprule
Optimizer & OA (\%) & AA (\%) & Kappa & Time (s/epoch) \\
\midrule
AdamW    & 99.15 & 99.18 & 0.99 & 8.00  \\
AdamMini & 98.95 & 98.99 & 0.99 & 8.00 \\
SGD      & 98.82 & 98.83 & 0.99 & 8.00 \\
\bottomrule
\end{tabular*}
\end{table}

Epoch count is varied between 50 and 600 in Table~\ref{tab:epochs} with stage budgets scaled proportionally. In common pipelines, too few epochs lead to underfitting and too many risk overfitting; the observed plateau with a peak near 300 epochs indicates that staged spectra guide optimization toward a stable basin without intensive schedule tuning.

\begin{table}[t]
\caption{Performance comparison across different epochs.}
\label{tab:epochs}
\centering
\small
\begin{tabular*}{\textwidth}{@{\extracolsep{\fill}}lccc}
\toprule
Epoch & OA (\%) & AA (\%) & Kappa \\
\midrule
 50  & 98.67 & 98.67 & 0.99 \\
 100 & 98.80 & 98.88 & 0.99 \\
 200 & 98.87 & 98.93 & 0.99 \\
 300 & 99.15 & 99.18 & 0.99 \\
 400 & 98.98 & 98.98 & 0.99 \\
 500 & 98.86 & 98.94 & 0.99 \\
 600 & 99.00 & 99.00 & 0.99 \\
\bottomrule
\end{tabular*}
\end{table}

\begin{table}[t]
  \caption{\textcolor{blue}{Robustness results under different random seeds and train-test splits. S1-S6: Different Random Seeds (fixed 80\% split). D1-D6: Different Training Ratios (fixed seed=42) }}
  \label{tab:robust}
  \centering
  \small
  \setlength{\tabcolsep}{15.0pt}
  \begin{tabular}{llccc}
    \bottomrule
    \# & Settings & OA (\%) & AA (\%) & Kappa \\
    \midrule
    S1 & Seed = 1     & 98.87 & 98.85 & 0.98 \\
    S2 & Seed = 10    & 98.80 & 98.87 & 0.98 \\
    S3 & Seed = 42    & 98.96 & 98.99 & 0.99 \\
    S4 & Seed = 123   & 98.96 & 98.94 & 0.99 \\
    S5 & Seed = 666   & 98.83 & 98.92 & 0.98 \\
    S6 & Mean $\pm$ Std & $98.88\pm0.07$ & $98.91\pm0.06$ & $0.984\pm0.005$ \\
    \midrule
    D1 & 10\% Train & 69.44 & 70.78 & 0.69       \\
    D2 & 20\% Train & 87.21 & 87.61 & 0.87       \\
    D3 & 30\% Train & 94.19 & 94.15 & 0.94       \\
    D4 & 40\% Train & 97.34 & 97.37 & 0.97        \\
    D5 & 50\% Train & 98.34 & 98.30 & 0.98       \\
    D6 & Mean $\pm$ Std & $89.30\pm11.93$ & $89.64\pm11.35$ & $0.89\pm0.12$ \\
    \bottomrule
  \end{tabular}
\end{table}

\textcolor{blue}{\subsection{Stability Testing}}
\textcolor{blue}{To comprehensively evaluate the proposed SpectralTrain method, we conducted robustness and stability tests using the 3D-ConvSST network as a baseline. We designed two sets of repeated experiments under strict univariate control conditions:}

\textcolor{blue}{1. Five repeated experiments with different random seeds (1, 10, 42, 123, 666) under a fixed 80\%/20\% train-test split. The mean value and standard deviation of OA are reported to eliminate the randomness caused by network initialization and data augmentation.}

\textcolor{blue}{2. Five groups of experiments with varying training sample proportions (10\%, 20\%, 30\%, 40\%, 50\%) with a fixed random seed 42. This part validates the model’s performance sensitivity to dataset random division and its capability for limited labeled hyperspectral cloud pixel classification.
The detailed quantitative results are summarized in Table \ref{tab:robust}.}

\subsection{Cross-dataset Study}
Cross-dataset behavior and the relation to recent architectures are summarized in Tables~\ref{tab:ip}--\ref{tab:cp7}. On Indian Pines, several backbones paired with SpectralTrain reach OA around or above 98.7\%, whereas DSFormer trained with its default settings performs worse on this dataset. On Salinas-A, accuracy remains high despite larger spectral variability. On CloudPatch-7 with 462 bands, most models approach 99\% OA/AA with Kappa at or above 0.98. Instances where an earlier architecture with SpectralTrain exceeds DSFormer are better interpreted not as an absolute statement about model capacity but as evidence that modern designs may be over-optimized for full-spectrum-from-epoch-1 regimes; reshaping early optimization through an information-preserving spectral schedule reduces reliance on capacity alone. This interpretation supports the joint design of architectures and training schedules in hyperspectral learning.

\begin{table}[t]
\caption{Performance comparison on Indian Pines dataset.}
\label{tab:ip}
\centering
\small
\begin{tabular*}{\textwidth}{@{\extracolsep{\fill}}lccc}
\toprule
Model & OA (\%) & AA (\%) & Kappa \\
\midrule
ADGAN       & 97.37 & 94.59 & 0.97 \\
MetaFormer  & 94.34 & 93.31 & 0.94 \\
ConvNeXt-T  & 98.72 & 99.12 & 0.99 \\
SQS         & 94.02 & 97.30 & 0.94 \\ 
3D-ConvSST  & 90.95 & 84.56 & 0.91 \\
RCNN        & 98.33 & 98.71 & 0.98 \\
DSFormer    & 90.58 & 77.05 & 0.91 \\
\bottomrule
\end{tabular*}
\end{table}

\begin{table}[t]
\caption{Performance comparison on Salinas-A dataset.}
\label{tab:sa}
\centering
\small
\begin{tabular*}{\textwidth}{@{\extracolsep{\fill}}lccc}
\toprule
Model & OA (\%) & AA (\%) & Kappa \\
\midrule
ADGAN       & 92.69 & 94.23 & 0.93 \\
MetaFormer  & 94.10 & 94.22 & 0.94 \\
ConvNeXt-T  & 98.09 & 97.86 & 0.98 \\
SQS         & 95.02 & 95.37 & 0.95 \\
3D-ConvSST  & 97.35 & 96.65 & 0.97 \\
RCNN        & 98.09 & 98.02 & 0.98 \\
DSFormer    & 91.20 & 89.81 & 0.91 \\
\bottomrule
\end{tabular*}
\end{table}

\begin{table}[t]
\caption{Performance comparison on CloudPatch-7 dataset.}
\label{tab:cp7}
\centering
\small
\begin{tabular*}{\textwidth}{@{\extracolsep{\fill}}lccc}
\toprule
Model & OA (\%) & AA (\%) & Kappa \\
\midrule
ADGAN       & 98.28 & 98.10 & 0.98 \\
MetaFormer  & 98.95 & 98.94 & 0.99 \\
ConvNeXt-T  & 99.15 & 99.18 & 0.99 \\
SQS         & 98.17 & 98.11 & 0.98 \\
3D-ConvSST  & 99.02 & 99.04 & 0.99 \\
RCNN        & 99.44 & 99.42 & 0.99 \\
DSFormer    & 93.75 & 93.81 & 0.94 \\
\bottomrule
\end{tabular*}
\end{table}

\begin{figure*}[]
  \centering
  \begin{minipage}[t]{0.95\textwidth}
    \centering
    \includegraphics[width=\linewidth]{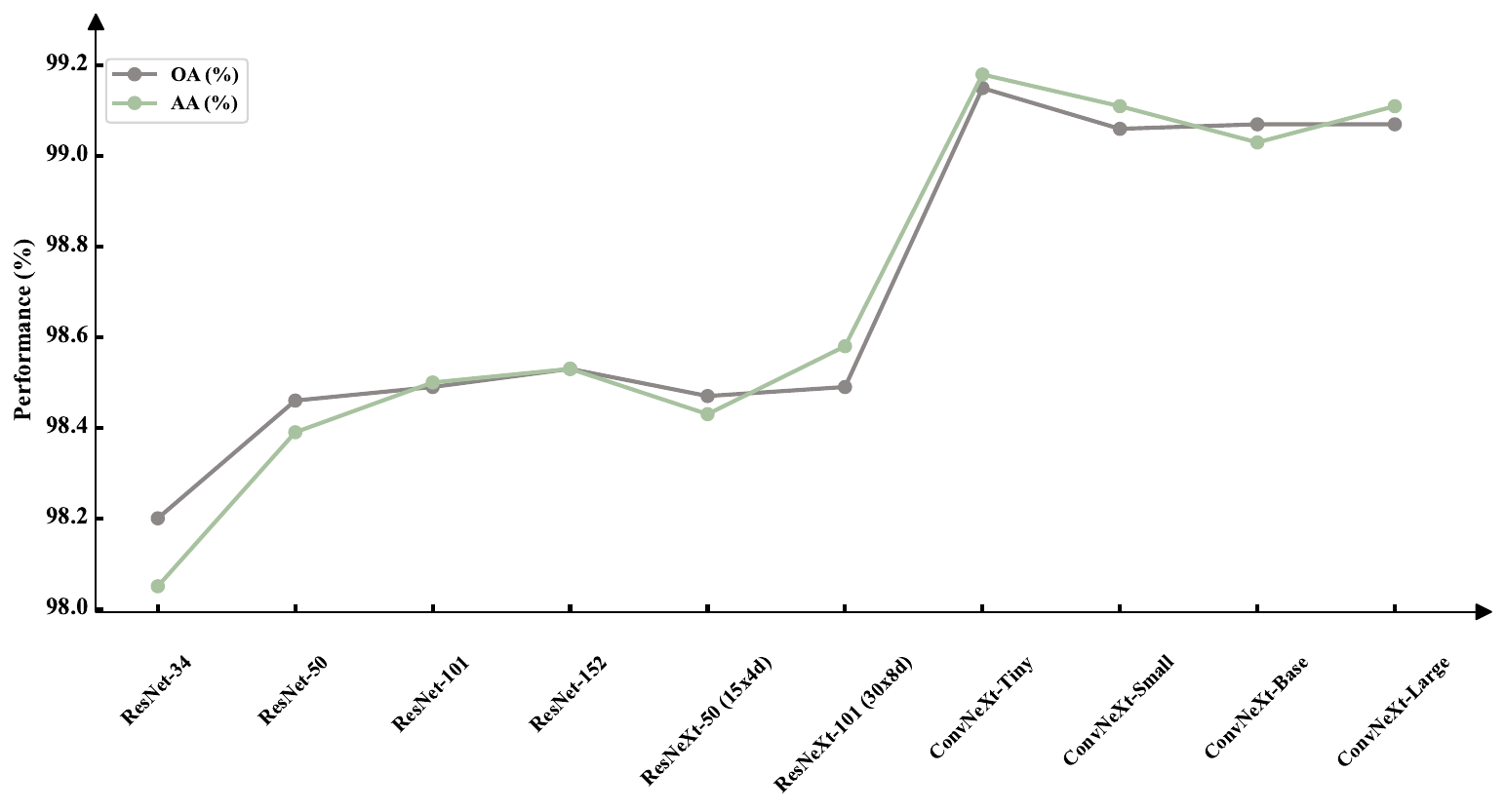}
    \vspace{-0.3em}
    \subcaption{OA and AA across different backbone scales}
  \end{minipage}

  \begin{minipage}[t]{0.95\textwidth}
    \centering
    \includegraphics[width=\linewidth]{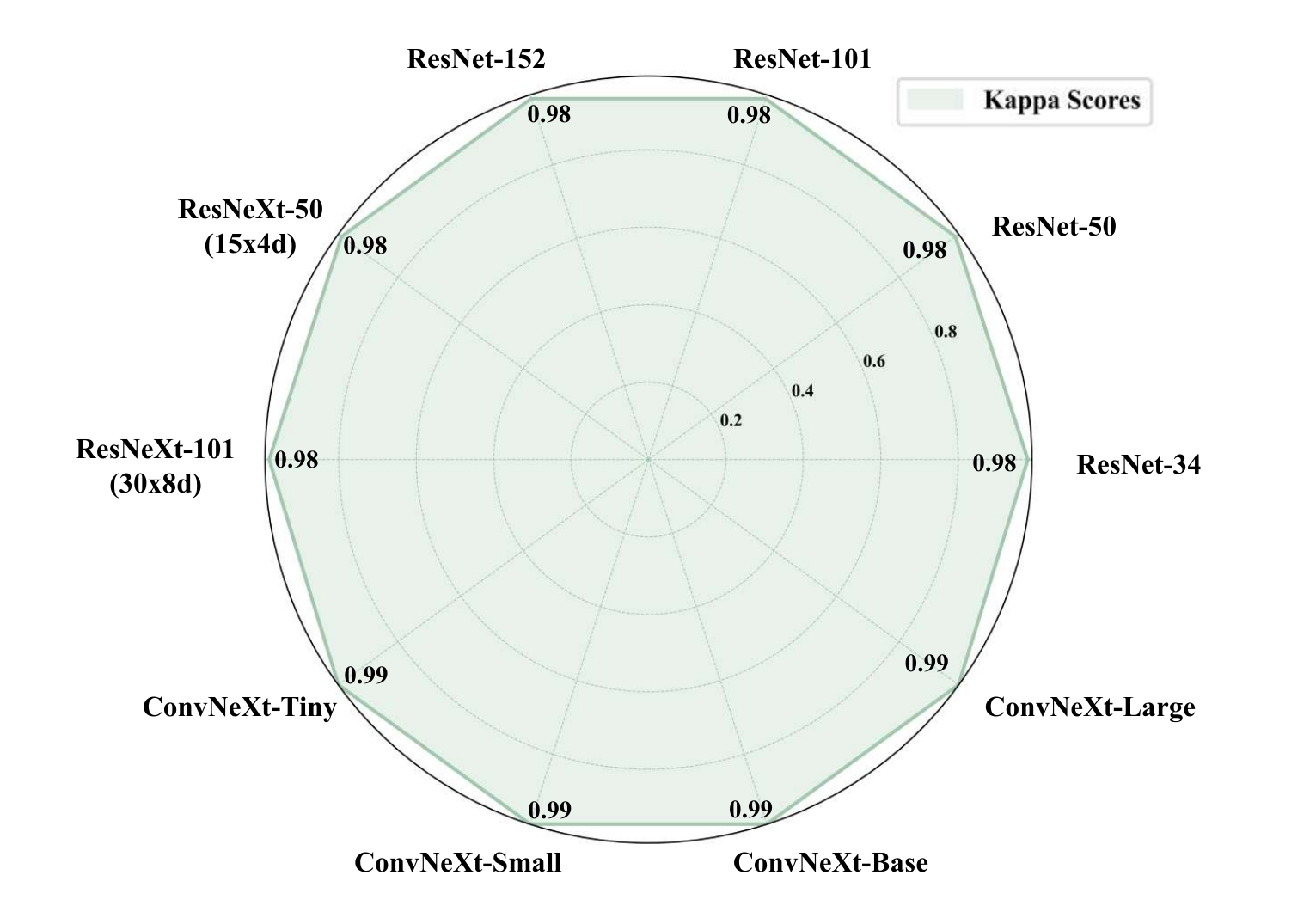}
    \vspace{-0.3em}
    \subcaption{\textcolor{blue}{Cohen's Kappa across ResNet and ConvNeXt variants}}
  \end{minipage}
  \caption{Backbone-scale performance under SpectralTrain.}
  \label{fig:backbone-scale}
\end{figure*}

The backbone-level trends are summarized in Figure~\ref{fig:backbone-scale}. Figure~\ref{fig:backbone-scale}(a) reports overall accuracy (OA) and average accuracy (AA) for different ResNet and ConvNeXt variants, while Figure~\ref{fig:backbone-scale}(b) reports Cohen's Kappa for the same variants. In conventional hyperspectral image classification, model depth and width often exert a pronounced effect on performance because larger capacity and receptive fields typically improve feature expressiveness. Under the proposed spectral curriculum, the dispersion across depth scales is small: OA and AA remain consistent from shallow to deep variants (for example, ResNet-34 to ResNet-152 and ConvNeXt-Tiny to ConvNeXt-Large), and Kappa values cluster around 0.99. These observations indicate that aligning the training schedule with spectral structure reduces the usual sensitivity to backbone depth and receptive-field configuration, which is consistent with the hypothesis that early spectral staging mitigates over-dependence on capacity.

\subsection{Hyperspectral Cloud Scene Understanding}

Inspired by the human practice of ``reading clouds to predict weather,''
this study uses CloudPatch-7 as a hyperspectral dataset for cloud-type
classification under extreme-weather-relevant conditions. This setting
connects atmospheric remote sensing with hyperspectral representation
learning and supports the analysis of cloud morphology through combined
spectral and spatial information.

The CloudPatch-7 dataset consists of 380 curated HSI patches ($50\times 50$ pixels) extracted from 28 full-scene captures, spanning a continuous spectral range of 462 bands from 400 to 1000 nm. The patches are classified into seven meteorological categories: dense dark cumuliform clouds, dense bright cumuliform clouds, semi-transparent cumuliform clouds, dense cirroform clouds, semi-transparent cirroform clouds, clear sky with low aerosol scattering, and clear sky with moderate to high aerosol scattering. These labels represent key atmospheric features relevant to climate monitoring and extreme weather forecasting.
To assess the effectiveness of SpectralTrain in atmospheric scene understanding, we performed classification experiments on CloudPatch-7 using seven representative model architectures across three structural paradigms: classical CNNs (ResNet-34, ConvNeXt-T), hybrid convolutional-transformer models (MetaFormer, ADGAN), and spectral-aware models designed for HSI processing (3D-ConvSST, SQS, DSFormer). Despite significant architectural differences, SpectralTrain consistently enabled all models to achieve high classification accuracy. Notably, even under limited computational resources and without pretraining, SpectralTrain preserved fine-grained spectral distinctions, enabling the separation of subtle atmospheric classes, such as semi-transparent cirroform clouds and high-scattering clear skies.

The robustness of these results underscores the practical applicability of SpectralTrain for cloud-type classification tasks in environmental monitoring. Accurate recognition of dense cumuliform or cirroform structures, for example, facilitates more precise tracking of convective storms and jet stream activity—key indicators of extreme weather events. Additionally, the ability to differentiate aerosol scattering levels from clear-sky signatures presents opportunities for high-resolution air quality monitoring.

Beyond the current scope, the consistently high performance observed across diverse models suggests that SpectralTrain may serve as a foundation for large-scale cloud analytics pipelines. Future work may explore integrating this framework with satellite-based imaging modalities and spatiotemporal modeling techniques to support long-term climate observation and early warning systems. Moreover, extending the CloudPatch series to include multi-temporal cloud dynamics or joint spectral–thermal modalities could further broaden the impact of hyperspectral learning in meteorological research.

\section{Discussion}\label{sec5}

This study reframes hyperspectral image classification as a setting where
training efficiency must be considered alongside architectural design.
\textcolor{blue}{The motivation figure
(Figure~\ref{fig:motivation}) highlights band-localized cues, contiguous
instrument responses, and bandwidth-bound training pipelines in
hyperspectral data. These properties limit the direct transfer of efficiency
strategies developed for RGB images.}
SpectralTrain organizes optimization along the spectral axis. Training begins
with an information-preserving compression and progressively restores full
spectral complexity as learning stabilizes.
\textcolor{blue}{This process aligns the evolution of input complexity with
the spectral structure of hyperspectral classification.}

EfficientTrain++ \cite{8} showed that a resolution-based curriculum can
reduce early training cost for RGB recognition.
\textcolor{blue}{SpectralTrain extends this curriculum principle to spectral
complexity by progressively increasing the information presented along the
spectral dimension. This distinction is relevant because spectral aliasing,
sensor response shifts, and bandwidth demands follow the
wavelength-dependent organization of hyperspectral data. Spatial and
spectral curricula address complementary dimensions of input complexity,
and their integration provides a natural direction for future research.}

Under the proposed schedule, several medium capacity or earlier backbones
match or surpass DSFormer \cite{30} on multiple datasets.
\textcolor{blue}{The comparison with conventional training \cite{31} further
indicates that architectural rankings can vary with the training protocol.
These results suggest that performance differences attributed to model
capacity may also depend on the spectral complexity presented during
optimization.}
Aligning the schedule with spectral structure attenuates sensitivity to
depth, receptive field, and token mixers, indicating that peak accuracy and
efficiency in hyperspectral classification are best pursued through
co-design of architectures and training strategies. Reporting both a
conventional and a curriculum-aligned protocol, as in
Table~\ref{tab:traincompare}, provides a fairer basis for architectural
comparison.

\textcolor{blue}{\textit{Limitations.}
The current evaluation covers hyperspectral classification on three datasets
under one hardware environment. The magnitude of the acceleration may vary
when spectral dimensionality is modest or backbone computation dominates
data transfer. This dependence follows from the operating mechanism of
SpectralTrain, which mainly reduces spectral input complexity and associated
system costs during the early training stages. Broader evaluations across
tasks, sensing platforms, and hardware configurations will further
characterize its applicable operating range.}

\textcolor{blue}{The experiments with principal component analysis, uniform
manifold approximation and projection, and independent component analysis
also clarify the role of the compression operator. The three reducers
produce comparable overall accuracy and average accuracy when the curriculum
boundaries, learning rate schedule, and computational budgets remain fixed,
as reported in Table~\ref{tab:dr}. This consistency supports the applicability
of staged spectral restoration across the evaluated compression operators
and identifies the progressive control of spectral complexity as the shared
mechanism underlying the efficiency gains.}

Two forward directions follow. For dense prediction, the same spectral
curriculum can be applied at the patch or pixel level with segmentation
encoders and decoders.
\textcolor{blue}{This direction is compatible with segmentation pretraining
\cite{41,42} and can examine whether curriculum-aligned optimization
preserves boundary information without increasing inference cost.}
For climate applications, multimodal cloud analysis that fuses hyperspectral,
thermal, and active sensors such as radar or LiDAR would benefit from a
modality-aware curriculum in which early phases operate on a low-cost
spectral subspace with down-weighted auxiliary streams, followed by
progressive increases in spectral rank and modality weights.
\textcolor{blue}{Such schedules could retain the system level advantages of
lower data transfer, smaller activation storage, and larger feasible batch
sizes that support the observed $2$--$7\times$ acceleration.}

\section{Conclusion}\label{sec6}
SpectralTrain is a training schedule for hyperspectral classification that starts from an information-preserving compressed spectrum and progressively restores full spectral complexity. Across Indian Pines, Salinas-A, and CloudPatch-7, the approach reduces training time by $2$–$7\times$ at comparable accuracy under diverse backbones, optimizers, losses, and epoch budgets. The findings indicate that efficiency in hyperspectral learning is governed as much by the schedule as by the architecture; co-designing the two yields more reliable speed–accuracy trade-offs than optimizing either in isolation. Future work will extend the curriculum to dense prediction with segmentation backbones and to multimodal cloud analysis in which hyperspectral data are fused with thermal and active sensors, aiming to retain the same resource profile while improving scene-level understanding.
\backmatter
\section*{Conflict of interest}
The authors declare that they have no conflicts of interest or competing interests.

\section*{Compliance with ethical standards}
This study did not involve human participants or animals. It relies solely on publicly available benchmark datasets and open-source code; therefore, approval from an institutional review board or ethics committee was not required. All authors have read and approved the final manuscript and agree with its submission.

\section*{Informed consent}
Not applicable. The study uses only publicly available, de-identified datasets; any consent was obtained by the original data providers in accordance with their policies.

\section*{Funding Information}
This work received no specific grant from any funding agency in the public, commercial, or not-for-profit sectors.

\section*{Data availability}
The experimental data and the codes that support the findings of this study are available in Github with the identifier \url{https://github.com/mh-zhou/SpectralTrain}.

\begin{appendices}

\section{CloudPatch\texorpdfstring{-}{-}7: Dataset Summary and Supplementary Visualizations}\label{secA1}

CloudPatch\mbox{-}7 is a hyperspectral benchmark for cloud-type classification under extreme-weather–relevant conditions. It comprises 380 manually annotated image patches of size $50\times50$ pixels extracted from 28 larger parent scenes (approximately $5000\times1500$ pixels each). Seven atmospheric categories are defined: dense dark cumuliform (46), dense bright cumuliform (77), semi-transparent cumuliform (72), dense cirroform (25), semi-transparent cirroform (28), clear sky with low aerosol scattering (68), and clear sky with moderate-to-high aerosol scattering (64). Across all patches there are $380\times50\times50=950{,}000$ labeled pixels, each with 462 spectral bands within 400–1000\,nm. The figures below summarize class balance and spectral statistics from code-generated analyses.

\begin{figure}[H]
  \centering
  \includegraphics[width=0.85\textwidth]{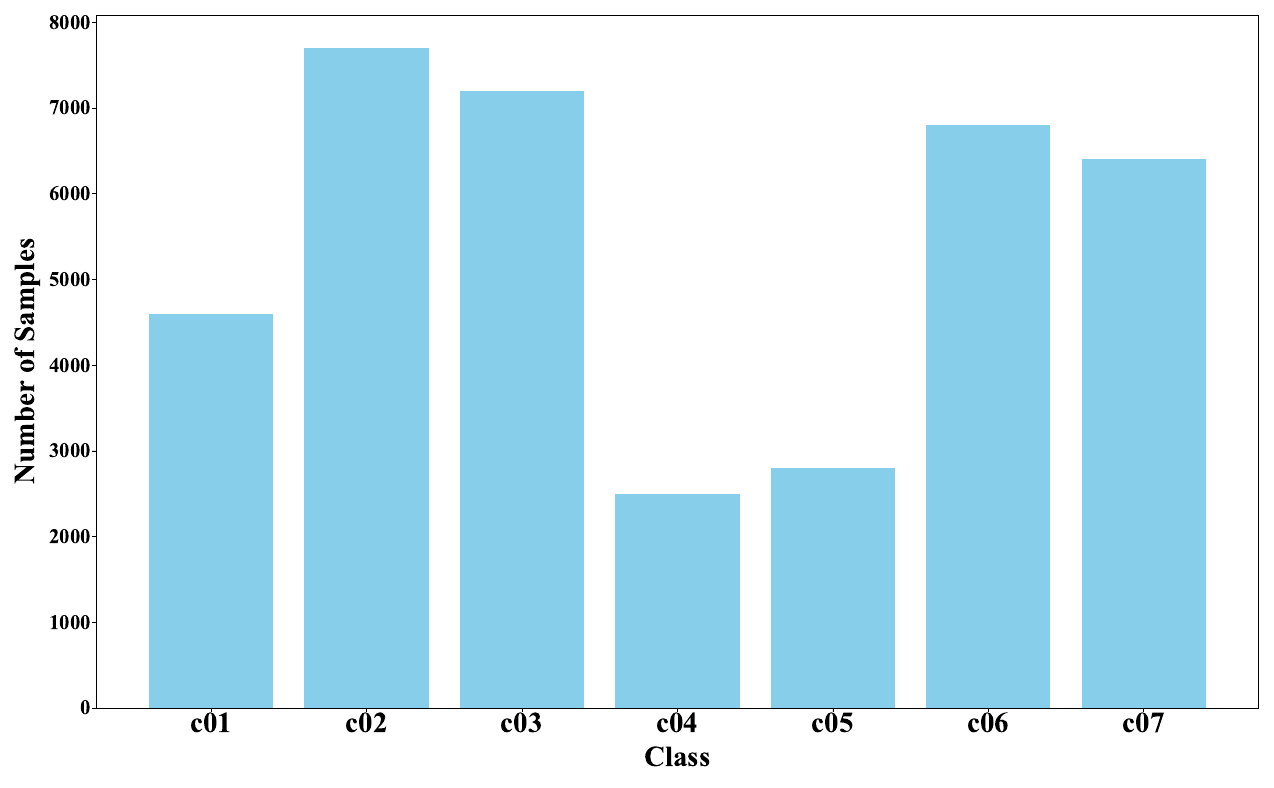}
  \caption{Supplementary Figure~S1. Class distribution in CloudPatch\mbox{-}7. Bars show per-class sample counts (c01–c07), revealing moderate imbalance.}
  \label{fig:supp-classdist}
\end{figure}

\begin{figure}[H]
  \centering
  \includegraphics[width=\textwidth]{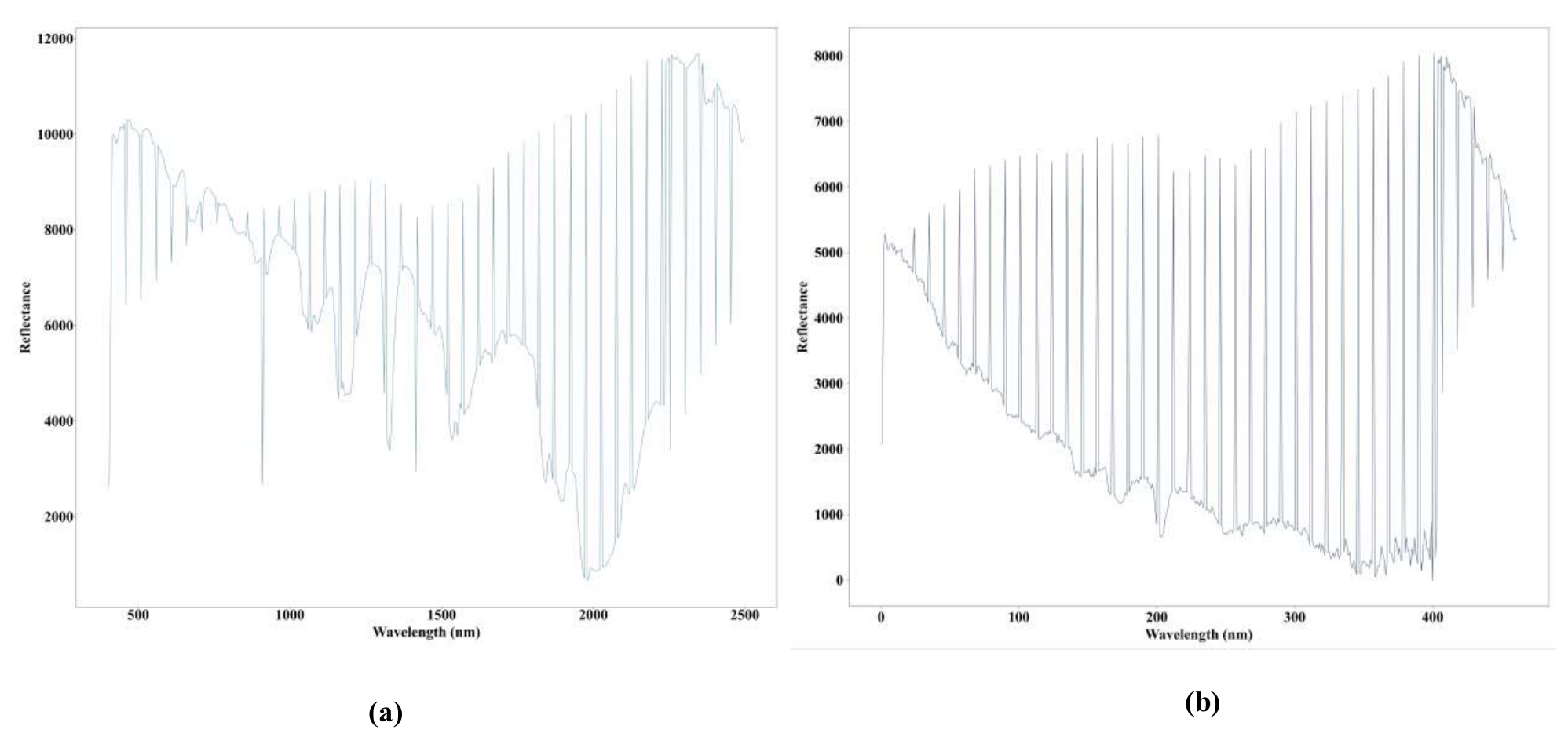}
  \caption{Supplementary Figure~S2. Spectral reflectance curves. Left (a): class-wise mean spectra; Right (b): a representative single-pixel spectrum. Axes are reflectance (vertical) and wavelength index (horizontal).}
  \label{fig:supp-mean-single}
\end{figure}

\begin{figure}[H]
  \centering
  \includegraphics[width=0.85\textwidth]{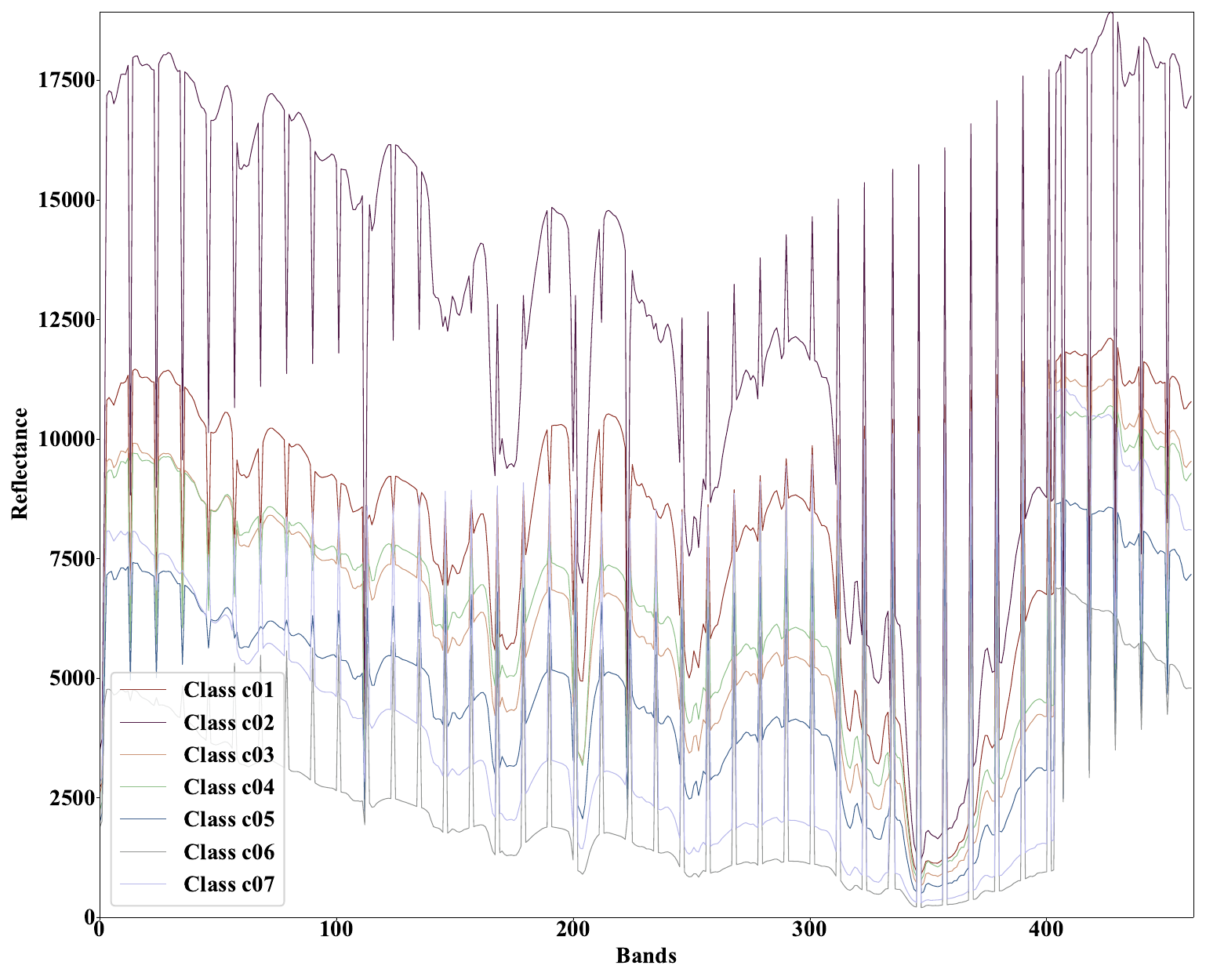}
  \caption{Supplementary Figure~S3. Average reflectance spectra for the seven classes (c01–c07). Distinct spectral shapes and band-localized differences provide discriminative cues.}
  \label{fig:supp-class-means}
\end{figure}

\begin{figure}[H]
  \centering
  \includegraphics[width=0.85\textwidth]{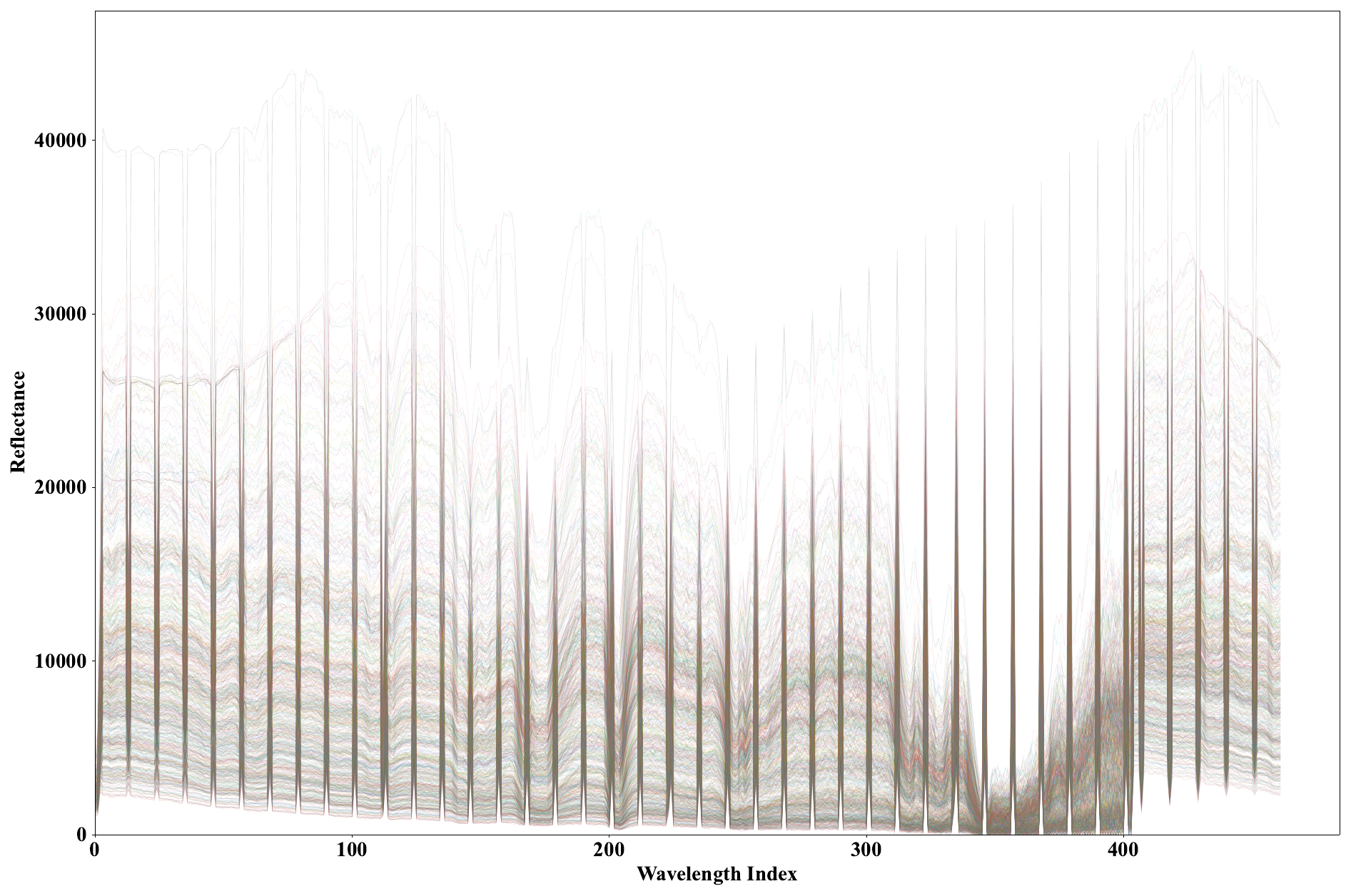}
  \caption{Supplementary Figure~S4. Dense overlay of per-sample spectra. Overlaid curves emphasize intra-class variability and contiguous, instrument-governed spectral ordering.}
  \label{fig:supp-dense}
\end{figure}

\section{Mathematical Justification for PCA-First Sampling
versus Low-Frequency Cropping}
\label{supp:math}

\begingroup
\color{blue}

Let a hyperspectral image be represented by
$I:\Omega\times\Lambda\rightarrow\mathbb{R}$, where
$\Omega\subset\mathbb{R}^{2}$ is the spatial domain and
$\Lambda=\{\lambda_1,\ldots,\lambda_D\}$ contains the $D$ spectral
bands. For a pixel $p\in\Omega$, its spectrum is
\[
s(p)
=
\bigl[
I(p,\lambda_1),\ldots,I(p,\lambda_D)
\bigr]^{\top}
\in\mathbb{R}^{D}.
\]
Stacking the spectra of $M$ pixels gives
$X\in\mathbb{R}^{M\times D}$. The notation follows the main theoretical
analysis. In particular, $D$ denotes the original spectral dimension,
while $N$ denotes the number of curriculum stages.

\subsection*{B.1\; Low-Frequency Cropping}

For each spectral band $\lambda_j$, ideal low-frequency cropping is
defined by
\[
\widehat{I'}(u,v,\lambda_j)
=
\widehat{I}(u,v,\lambda_j)H(u,v),
\qquad
H(u,v)
=
\mathbf{1}
\left\{
\sqrt{u^2+v^2}\leq f_c
\right\}.
\]
The convolution theorem gives
\[
I'(\cdot,\cdot,\lambda_j)
=
I(\cdot,\cdot,\lambda_j)*h,
\qquad
h=\mathcal{F}^{-1}\{H\}.
\]

\begin{theorem}
\label{thm:supp_lfc}
Let $\mathcal{T}_Hf=f*h$, where the Fourier multiplier satisfies
$|H(u,v)|\leq1$. For any $f,g\in L^2(\Omega)$,
\begin{equation}
\|\mathcal{T}_Hf\|_{L^2}
\leq
\|f\|_{L^2},
\label{eq:supp_lfc_contraction}
\end{equation}
and
\begin{equation}
\|\mathcal{T}_Hf-\mathcal{T}_Hg\|_{L^2}
\leq
\|f-g\|_{L^2}.
\label{eq:supp_lfc_pair}
\end{equation}
When $H$ is the indicator of the retained frequency region, the discarded
energy satisfies
\begin{equation}
\|f\|_{L^2}^{2}
-
\|\mathcal{T}_Hf\|_{L^2}^{2}
=
\int_{\{(u,v):H(u,v)=0\}}
|\widehat{f}(u,v)|^2\,du\,dv.
\label{eq:supp_lfc_energy}
\end{equation}
The original signal is recoverable from $\mathcal{T}_Hf$ only when the
right-hand side of Eq.~\eqref{eq:supp_lfc_energy} is zero.
\end{theorem}

\begin{proof}
Parseval's identity and the convolution theorem give
\begin{align}
\|\mathcal{T}_Hf\|_{L^2}^{2}
&=
\|\widehat{f}H\|_{L^2}^{2} \\
&=
\int
|\widehat{f}(u,v)|^2
|H(u,v)|^2\,du\,dv \\
&\leq
\int
|\widehat{f}(u,v)|^2\,du\,dv \\
&=
\|f\|_{L^2}^{2},
\end{align}
which proves Eq.~\eqref{eq:supp_lfc_contraction}. Applying the same
argument to $f-g$ gives Eq.~\eqref{eq:supp_lfc_pair}.

For an indicator multiplier, $|H|^2=H$. Therefore,
\begin{align}
\|f\|_{L^2}^{2}
-
\|\mathcal{T}_Hf\|_{L^2}^{2}
&=
\int
|\widehat{f}(u,v)|^2
\bigl(1-H(u,v)\bigr)\,du\,dv \\
&=
\int_{\{(u,v):H(u,v)=0\}}
|\widehat{f}(u,v)|^2\,du\,dv.
\end{align}
Any nonzero energy outside the retained frequency region is absent from
the cropped result, which proves Eq.~\eqref{eq:supp_lfc_energy} and the
recoverability statement.
\end{proof}

\begin{remark}
The inequality in Eq.~\eqref{eq:supp_lfc_pair} is strict when
$\widehat{f}-\widehat{g}$ contains nonzero energy in the removed frequency
region. For hyperspectral data, the spatial filter is applied to every
band image. At a pixel $p$, the transformed spectrum satisfies
\begin{equation}
s'(p)
=
\int_{\Omega}
s(q)h(p-q)\,dq.
\label{eq:supp_spectral_mixing}
\end{equation}
When a practical low-pass kernel satisfies $h\geq0$ and
$\int_{\Omega}h=1$, Eq.~\eqref{eq:supp_spectral_mixing} is a weighted
average of neighboring spectra. Spectral contrast can consequently decrease
near class boundaries whose filtering support contains multiple classes.
\end{remark}

\subsection*{B.2\; Spatial Locality of PCA Projection}

Let
\[
\widetilde{X}
=
X-\mathbf{1}_{M}\mu^{\top}
\]
be the centered spectral matrix, and let
\[
E_K
=
[e_1,\ldots,e_K]
\in
\mathbb{R}^{D\times K}
\]
contain the leading principal component analysis (PCA) directions. The
retained PCA coordinates are
\[
Y_K
=
\widetilde{X}E_K
\in
\mathbb{R}^{M\times K}.
\]
The optimal rank-$K$ approximation error and its monotonic decrease with
$K$ are established by Theorem~\ref{thm:nested_pca}. The following result
characterizes the spatial locality of the PCA operation.

\begin{theorem}
\label{thm:supp_commutation}
Let $G\in\mathbb{R}^{M'\times M}$ represent a linear spatial operator
applied identically to every spectral band. Then
\begin{equation}
G\bigl(\widetilde{X}E_K\bigr)
=
\bigl(G\widetilde{X}\bigr)E_K.
\label{eq:supp_commutation}
\end{equation}
Consequently, PCA projection introduces no cross-pixel aggregation by
itself. Spatial mixing is determined by the operator $G$.
\end{theorem}

\begin{proof}
Matrix associativity directly gives
\[
G\bigl(\widetilde{X}E_K\bigr)
=
\bigl(G\widetilde{X}\bigr)E_K.
\]
The multiplication by $E_K$ acts independently on each row of
$\widetilde{X}$, where each row corresponds to one spatial position.
Therefore, PCA projection changes the spectral coordinates at each
position without aggregating information across positions.
\end{proof}

\begin{remark}
Equation~\eqref{eq:supp_commutation} permits PCA projection and a shared
linear anti-aliasing operator to be evaluated in either order. PCA removes
directions along the spectral axis, while the spatial operator determines
whether neighboring pixels are averaged. This distinction complements the
spectral approximation result in Theorem~\ref{thm:nested_pca}.
\end{remark}

\subsection*{B.3\; Nyquist-Consistent Sampling of PCA Images}

Let
\[
y_k(x,y)
=
e_k^{\top}\widetilde{s}(x,y)
\]
denote the $k$th principal component image. After applying a normalized
low-pass filter $g$, sampling by an integer factor $d\geq2$ gives
\begin{equation}
y_{k,d}[m,n]
=
(y_k*g)(md,nd).
\label{eq:supp_pc_samples}
\end{equation}

\begin{theorem}
\label{thm:supp_nyquist}
Assume that the Fourier transform of $y_k*g$ vanishes whenever
\[
|\omega_x|>\frac{\pi}{d}
\qquad\text{or}\qquad
|\omega_y|>\frac{\pi}{d}.
\]
Then the filtered principal component image is reconstructed from
Eq.~\eqref{eq:supp_pc_samples} as
\begin{equation}
(y_k*g)(x,y)
=
\sum_{m,n\in\mathbb{Z}}
y_{k,d}[m,n]\,
\operatorname{sinc}
\left(
\frac{x}{d}-m
\right)
\operatorname{sinc}
\left(
\frac{y}{d}-n
\right).
\label{eq:supp_pc_interpolation}
\end{equation}
Applying Eq.~\eqref{eq:supp_pc_interpolation} independently to the first
$K$ principal component images reconstructs the retained spectral field
as
\begin{equation}
\widehat{s}_K(x,y)
=
\mu
+
E_K
\begin{bmatrix}
(y_1*g)(x,y)\\
\vdots\\
(y_K*g)(x,y)
\end{bmatrix}.
\label{eq:supp_spectral_reconstruction}
\end{equation}
The sampling and interpolation operations introduce no cross-component
mixing.
\end{theorem}

\begin{proof}
The assumed frequency support satisfies the two-dimensional
Shannon--Nyquist sampling criterion for interval $d$. Separable sinc
interpolation along the two spatial axes gives
Eq.~\eqref{eq:supp_pc_interpolation}. Applying the same reconstruction
independently to each of the first $K$ principal component images recovers
their coordinate vector. Multiplication by $E_K$ and restoration of the
spectral mean then give Eq.~\eqref{eq:supp_spectral_reconstruction}.
Because each reconstructed coordinate depends only on samples from the
same coordinate, no cross-component mixing is introduced.
\end{proof}

\begin{remark}
Theorem~\ref{thm:supp_nyquist} concerns the retained PCA representation
under ideal band-limited sampling. It does not recover directions excluded
from $\operatorname{span}(E_K)$. The expected residual energy of these
directions is quantified by Eq.~\eqref{eq:pca_expected_tail}. Practical
interpolation kernels can additionally introduce a spatial approximation
error when the ideal support condition is only approximately satisfied.
\end{remark}

\subsection*{B.4\; Implication for SpectralTrain}

\begin{remark}
The preceding results separate the spectral and spatial transformations
used by SpectralTrain. PCA operates independently at each spatial position
and removes spectral directions according to the covariance structure.
Spatial reduction operates on the retained principal component images and
requires anti-aliasing before sampling.

At curriculum stage $i$, the transformed input can be represented as
\begin{equation}
X_i
=
\mathcal{D}_{B_i}
G_i
\left(
\widetilde{X}E_{k_i}
\right),
\label{eq:supp_stage_input}
\end{equation}
where $G_i$ is the spatial anti-aliasing operator and
$\mathcal{D}_{B_i}$ samples the retained component images onto the
$B_i\times B_i$ grid. Applying PCA before GPU transfer allows the stage
to transfer and process only the active $k_i$ principal component images.
The corresponding spectral approximation error and input-dependent cost
are quantified by Theorems~\ref{thm:nested_pca} and
\ref{thm:input_cost}, respectively.
\end{remark}

\endgroup

\section{Explanatory Analysis under a Linear Surrogate}
\label{supp:math_c_step}

{\color{blue}

This supplement develops additional results under a linear prediction model
with squared loss. It complements the exact input-transformation results
and the conditional deep-network discrepancy bounds in
Section~\ref{subsec:theory}. The conclusions in this section apply under
the assumptions stated in each theorem. They do not provide global
convergence or unchanged-accuracy guarantees for the deep non-convex
architectures evaluated in the experiments.

We adopt the notation from Section~\ref{subsec:theory}. In particular,
$\widetilde{x}\in\mathbb{R}^{D}$ is centered,
\[
C
=
\mathbb{E}
\left[
\widetilde{x}\widetilde{x}^{\top}
\right]
=
E\Lambda E^{\top},
\]
and
\[
P_K=E_KE_K^{\top}
\]
is the rank-$K$ PCA projector.

\begin{theorem}
\label{thm:supp_linear_bias}
Assume that the response follows
\begin{equation}
y
=
\langle w^{\star},\widetilde{x}\rangle+\xi,
\label{eq:supp_linear_model}
\end{equation}
where
\[
\mathbb{E}[\xi\mid\widetilde{x}]=0.
\]
Let
\[
w_K=P_Kw^{\star}
\]
be the projection of the full linear predictor onto the leading
$K$ PCA directions. For the squared prediction risk
\[
R(w)
=
\mathbb{E}
\left[
\bigl(
\langle w,\widetilde{x}\rangle-y
\bigr)^2
\right],
\]
the approximation cost is
\begin{equation}
R(w_K)-R(w^{\star})
=
\sum_{j=K+1}^{D}
\lambda_j
\bigl(e_j^{\top}w^{\star}\bigr)^2.
\label{eq:supp_linear_bias}
\end{equation}
\end{theorem}

\begin{proof}
Substituting Eq.~\eqref{eq:supp_linear_model} gives
\begin{align}
R(w_K)
&=
\mathbb{E}
\left[
\left(
\langle w_K-w^{\star},\widetilde{x}\rangle-\xi
\right)^2
\right] \\
&=
\mathbb{E}
\left[
\langle w_K-w^{\star},\widetilde{x}\rangle^2
\right]
+
\mathbb{E}[\xi^2],
\end{align}
because
\[
\mathbb{E}
\left[
\xi
\langle w_K-w^{\star},\widetilde{x}\rangle
\right]
=
0.
\]
Similarly,
\[
R(w^{\star})=\mathbb{E}[\xi^2].
\]
Therefore,
\begin{align}
R(w_K)-R(w^{\star})
&=
(w_K-w^{\star})^{\top}
C
(w_K-w^{\star}) \\
&=
\bigl((I-P_K)w^{\star}\bigr)^{\top}
C
\bigl((I-P_K)w^{\star}\bigr).
\end{align}
Expanding $w^{\star}$ in the eigenbasis of $C$ yields
\[
(I-P_K)w^{\star}
=
\sum_{j=K+1}^{D}
(e_j^{\top}w^{\star})e_j.
\]
Substitution gives Eq.~\eqref{eq:supp_linear_bias}.
\end{proof}

\begin{remark}
The approximation cost depends on the alignment between the predictive
direction $w^{\star}$ and the discarded eigendirections. A small covariance
tail alone does not guarantee a small task-specific error. The result
explains why rare classes associated with low-variance spectral directions
can remain difficult under aggressive compression. Later curriculum stages
reduce this approximation term by increasing $K$.
\end{remark}

\begin{theorem}
\label{thm:supp_conditioning}
Consider the ridge-regularized squared-loss objective in the leading
$K$ PCA coordinates
\begin{equation}
F_K(a)
=
\frac{1}{2}
\mathbb{E}
\left[
\left(
a^{\top}E_K^{\top}\widetilde{x}-y
\right)^2
\right]
+
\frac{\gamma}{2}\|a\|_2^2,
\qquad
\gamma>0.
\label{eq:supp_ridge_objective}
\end{equation}
Its Hessian is
\begin{equation}
\nabla^2F_K
=
\operatorname{diag}
(\lambda_1,\ldots,\lambda_K)
+
\gamma I_K.
\label{eq:supp_ridge_hessian}
\end{equation}
Consequently, $F_K$ is
$(\lambda_K+\gamma)$-strongly convex and
$(\lambda_1+\gamma)$-smooth. Its condition number is
\begin{equation}
\kappa_K
=
\frac{\lambda_1+\gamma}
{\lambda_K+\gamma}.
\label{eq:supp_condition_number}
\end{equation}
For $K\leq K'$,
\begin{equation}
\kappa_K\leq\kappa_{K'}.
\label{eq:supp_condition_monotone}
\end{equation}
\end{theorem}

\begin{proof}
The PCA-coordinate covariance is
\[
E_K^{\top}CE_K
=
\operatorname{diag}
(\lambda_1,\ldots,\lambda_K).
\]
Differentiating Eq.~\eqref{eq:supp_ridge_objective} twice gives
Eq.~\eqref{eq:supp_ridge_hessian}. Its largest and smallest eigenvalues
are $\lambda_1+\gamma$ and $\lambda_K+\gamma$, respectively. This proves
Eq.~\eqref{eq:supp_condition_number}.

Since the eigenvalues are ordered,
\[
\lambda_K\geq\lambda_{K'}
\]
whenever $K\leq K'$. Hence
\[
\lambda_K+\gamma
\geq
\lambda_{K'}+\gamma,
\]
which gives Eq.~\eqref{eq:supp_condition_monotone}.
\end{proof}

\begin{remark}
The theorem shows that the low-rank squared-loss surrogate has a smaller
or equal condition number than a higher-rank surrogate. This conclusion
depends on the linear model, squared loss, and ridge regularization. For
deep networks, the local loss geometry changes with the parameters, so the
result provides an explanatory analogue of the early-stage continuation
mechanism.
\end{remark}

\begin{theorem}
\label{thm:supp_gradient_variance}
Let
\[
z_K=E_K^{\top}\widetilde{x},
\]
and let the random part of a single-sample gradient have the form
\[
g_K=\psi(a;z_K,y)z_K.
\]
Assume that
\[
|\psi(a;z_K,y)|
\leq
C_{\psi}
\]
almost surely. Then
\begin{equation}
\operatorname{tr}
\left(
\operatorname{Cov}[g_K]
\right)
\leq
C_{\psi}^{2}
\sum_{j=1}^{K}\lambda_j.
\label{eq:supp_gradient_variance}
\end{equation}
For $K\leq K'$,
\begin{equation}
C_{\psi}^{2}
\sum_{j=1}^{K}\lambda_j
\leq
C_{\psi}^{2}
\sum_{j=1}^{K'}\lambda_j.
\label{eq:supp_gradient_variance_order}
\end{equation}
\end{theorem}

\begin{proof}
For any random vector $g$,
\[
\operatorname{tr}
\left(
\operatorname{Cov}[g]
\right)
\leq
\mathbb{E}\|g\|_2^2.
\]
Therefore,
\begin{align}
\operatorname{tr}
\left(
\operatorname{Cov}[g_K]
\right)
&\leq
\mathbb{E}
\left[
\psi(a;z_K,y)^2
\|z_K\|_2^2
\right] \\
&\leq
C_{\psi}^{2}
\mathbb{E}\|z_K\|_2^2.
\end{align}
The covariance of $z_K$ is
\[
\operatorname{diag}
(\lambda_1,\ldots,\lambda_K),
\]
so
\[
\mathbb{E}\|z_K\|_2^2
=
\sum_{j=1}^{K}\lambda_j.
\]
This proves Eq.~\eqref{eq:supp_gradient_variance}. The ordered partial sums
give Eq.~\eqref{eq:supp_gradient_variance_order}.
\end{proof}

\begin{remark}
The bound indicates that a lower-rank stage can reduce an upper bound on
the stochastic-gradient variance contributed by the input coordinates.
The result does not establish that every low-rank stage has lower observed
gradient variance, because the residual factor $\psi$ can change with the
model parameters and the active representation.
\end{remark}

\begin{theorem}
\label{thm:supp_time_condition}
Assume the conditions of
Theorem~\ref{thm:surrogate_continuation}. Let
\[
d_0
=
\|\Theta_0-\Theta^{\star}\|,
\qquad
0<q<1,
\]
and let the first curriculum stage use $t_1$ updates with approximation
bound $\varepsilon_K$. Define
\begin{equation}
d_1
=
q^{t_1}d_0
+
\frac{\varepsilon_K}{\mu}
\left(
1-q^{t_1}
\right).
\label{eq:supp_stage_distance}
\end{equation}
For a target distance $\tau>0$, define
\begin{equation}
t_{\mathrm{base}}
=
\left\lceil
\frac{\log(\tau/d_0)}
{\log q}
\right\rceil_{+},
\qquad
t_2
=
\left\lceil
\frac{\log(\tau/d_1)}
{\log q}
\right\rceil_{+},
\label{eq:supp_required_steps}
\end{equation}
where
\[
\lceil a\rceil_{+}
=
\max\{0,\lceil a\rceil\}.
\]
Let $c_K$ and $c_F$ denote the measured per-update costs of the early
configuration and the final configuration. The two-stage curriculum has
lower wall-clock cost than direct final-configuration optimization whenever
\begin{equation}
\frac{c_K}{c_F}
<
\frac{t_{\mathrm{base}}-t_2}{t_1}.
\label{eq:supp_time_condition}
\end{equation}
\end{theorem}

\begin{proof}
Direct optimization with the final configuration requires at most
$t_{\mathrm{base}}$ updates under the assumed contraction factor. Its cost
is
\[
C_{\mathrm{base}}
=
t_{\mathrm{base}}c_F.
\]
After the early stage, Eq.~\eqref{eq:supp_stage_distance} bounds the
distance to the surrogate minimizer. The final stage then requires at most
$t_2$ updates. The curriculum cost is
\[
C_{\mathrm{cur}}
=
t_1c_K+t_2c_F.
\]
The inequality
\[
C_{\mathrm{cur}}<C_{\mathrm{base}}
\]
is equivalent to
\[
t_1c_K
<
(t_{\mathrm{base}}-t_2)c_F.
\]
Dividing by $t_1c_F$ proves Eq.~\eqref{eq:supp_time_condition}.
\end{proof}

\begin{remark}
The theorem separates the optimization benefit from the systems cost.
The early stage is useful when it reduces the number of required
final-configuration updates sufficiently to offset its own cost. The
input-dependent component of $c_K/c_F$ is characterized by the main-text
cost theorem. Actual wall-clock values also depend on data loading,
backbone computation, memory access, and hardware utilization. The reported
deep-network speedups are therefore empirical results.
\end{remark}

\begin{remark}
The supplementary analysis establishes three mechanisms within the stated
linear-surrogate setting. Spectral truncation introduces a task-dependent
bias determined by discarded predictive directions. Lower-rank ridge
objectives can have more favorable conditioning and a smaller
stochastic-gradient variance bound. A curriculum reduces wall-clock cost
when the inexpensive early updates sufficiently reduce the subsequent
full-configuration optimization. These results explain the observed
training behavior under simplified assumptions and remain separate from
the empirical conclusions for deep non-convex networks.
\end{remark}

}
\end{appendices}

\bibliography{sn-bibliography}

\end{document}